\documentclass[sigconf,nonacm]{acmart}

\AtBeginDocument{%
  \providecommand\BibTeX{{%
    \normalfont B\kern-0.5em{\scshape i\kern-0.25em b}\kern-0.8em\TeX}}}

\setcopyright{acmcopyright}
\copyrightyear{2023}
\acmYear{2023}
\acmDOI{XXXXXXX.XXXXXXX}

\acmConference[CHI '24]{May 11--16,2024}{Honolulu, Hawai'i}

\acmSubmissionID{9694}

\usepackage[noabbrev,capitalise,nameinlink]{cleveref}
\usepackage{enumitem}
\usepackage{xspace}

\newcommand{\widesim}[2][1.5]{
  \mathrel{\overset{#2}{\scalebox{#1}[1]{$\sim$}}}
}

\begin{document}

\newcommand{\worldviewer}{\textsc{DiffusionWorldViewer}\xspace}
\newcommand{\subscript}[2]{$#1 _ #2$}
\definecolor{CaptaColor}{HTML}{1B75C3}
\definecolor{ArrowColor}{HTML}{e28743}
\definecolor{ContextColor}{HTML}{0F9873}
\definecolor{FunctionColor}{HTML}{5c5c5c}
\newcommand{\todo}[1]{{\textcolor{red}{[#1]}\normalfont}}
\newcommand{\capta}{\textcolor{CaptaColor}{\textsc{capta}}\xspace}
\newcommand{\arrows}{\textcolor{ArrowColor}{\textsc{arrows}}\xspace}
\newcommand{\arrow}{\textcolor{ArrowColor}{\textsc{arrow}}\xspace}
\newcommand{\context}{\textcolor{ContextColor}{\textsc{context}}\xspace}
\newcommand{\generate}{\textcolor{FunctionColor}{\texttt{generate}}\xspace}
\newcommand{\estimate}{\textcolor{FunctionColor}{\texttt{estimate}}\xspace}
\newcommand{\estimating}{\textcolor{FunctionColor}{\texttt{estimating}}\xspace}
\newcommand{\worldview}{\textsc{worldview}\xspace}
\newcommand{\concept}{\textcolor{ContextColor}{\textsc{concept}}\xspace}
\newcommand{\concepts}{\textcolor{ContextColor}{\textsc{concepts}}\xspace}

\title{%
 \worldviewer: Exposing and Broadening the Worldview Reflected by Generative Text-to-Image Models} 
\author{Zoe De Simone}
\orcid{0000-0001-9138-9362}
\affiliation{CSAIL
  \institution{MIT}
  \streetaddress{77 Massachusetts Ave, Cambridge, MA 02139}
  \city{Cambridge}
  \state{Massachusetts}
  \country{USA}
  \postcode{02139-4307}
}
\email{zoed@mit.edu}

\author{Angie Boggust}
\affiliation{CSAIL
  \institution{MIT}
  \streetaddress{77 Massachusetts Ave, Cambridge, MA 02139}
  \city{Cambridge}
  \state{Massachusetts}
  \country{USA}
  \postcode{02139-4307}}
\email{aboggust@csail.mit.edu}

\author{Arvind Satyanarayan}
\affiliation{CSAIL
  \institution{MIT}
  \streetaddress{77 Massachusetts Ave, Cambridge, MA 02139}
  \city{Cambridge}
  \state{Massachusetts}
  \country{USA}
  \postcode{02139-4307}}
\email{arvindsatya@mit.edu}

\author{Ashia Wilson}
\affiliation{LIDS
  \institution{MIT}
  \streetaddress{77 Massachusetts Ave, Cambridge, MA 02139}
  \city{Cambridge}
  \state{Massachusetts}
  \country{USA}
  \postcode{02139-4307}}
\email{ashia07@mit.edu}

\renewcommand{\shortauthors}{De Simone, et al.}

\begin{abstract}
Generative text-to-image (TTI) models produce high-quality images from short textual descriptions and are widely used in academic and creative domains. 
Like humans, TTI models have a \textit{worldview}, a conception of the world learned from their training data and task that influences the images they generate for a given prompt. 
However, the worldviews of TTI models are often hidden from users, making it challenging for users to build intuition about TTI outputs, and they are often misaligned with users' worldviews, resulting in output images that do not match user expectations. 
In response, we introduce \worldviewer\,---\,an interactive interface that exposes a TTI model's worldview across output demographics and provides editing tools for aligning output images with user perspectives. 
In a user study with 18 diverse TTI users, we find that \worldviewer helps users represent their varied viewpoints in generated images and challenge the limited worldview reflected in current TTI models.
\end{abstract}

\ccsdesc[500]{Human-centered computing~Interactive systems and tools}
\ccsdesc[500]{Computing methodologies~Computer vision}

\keywords{ worldviews, generative text to image models, HCI}

\begin{teaserfigure}
\includegraphics[width=\linewidth]{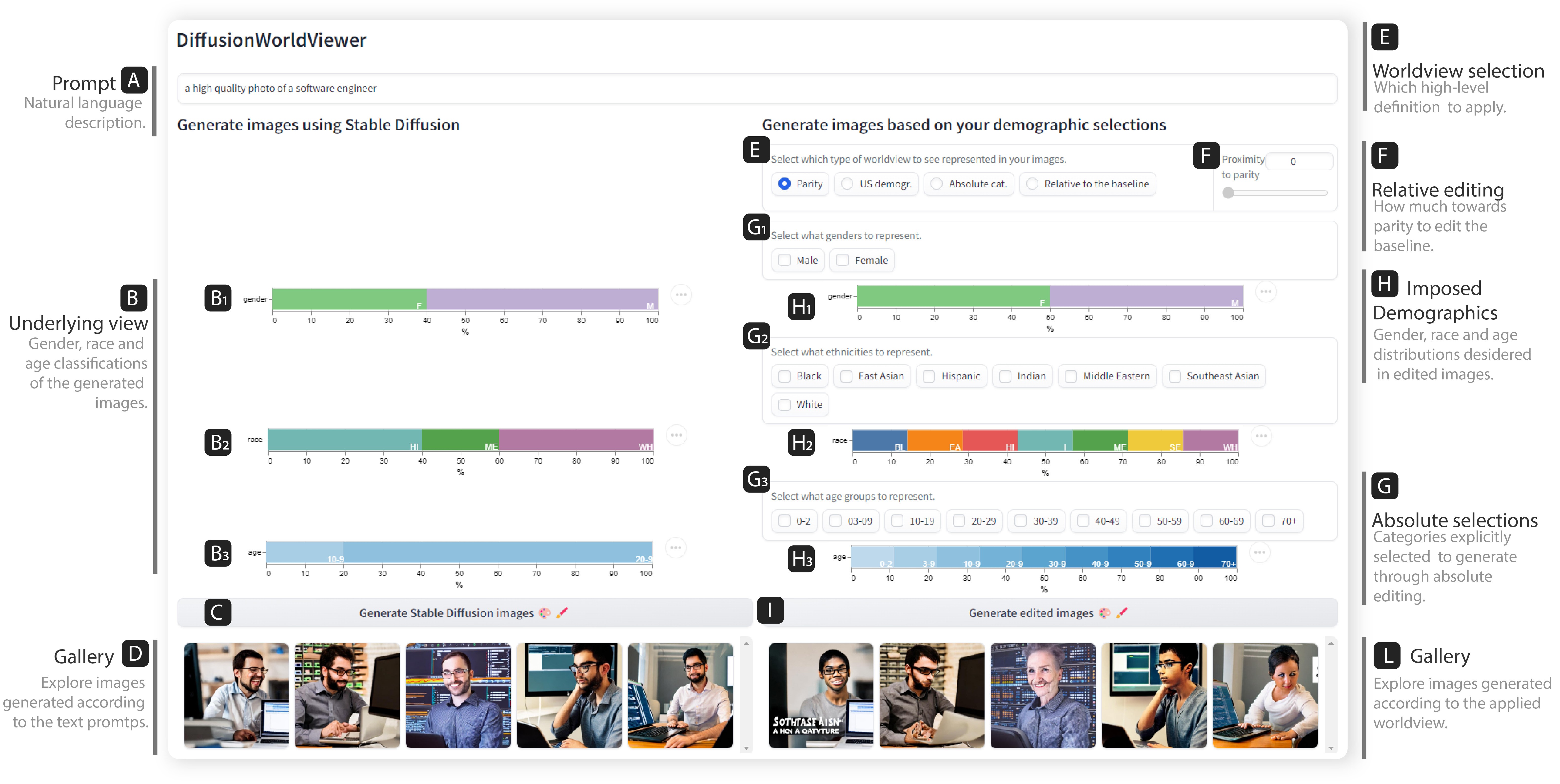}
\caption{\worldviewer allows users to explore underlying demographic distributions and worldviews of generative text-to-image models. 
Using \worldviewer, users can prompt a Stable Diffusion model (A) and visualize the gender (B1), race (B2), and age (B3) distributions of its generated images (C/D).
If the generated images conflict with a user's expectations, users can edit (E/F) the gender (G1/H1), race (G2/H2), and age (G3/H3) distributions which updates the generated images (I/L) to better reflect their worldview.}
\Description{DiffusionWorldViewer user interface with 2 column design. The page consists of a text box for users to type their prompts at the top where a user has typed in "a high quality photo of a software engineer" (labelled A), followed by a main area subdivided into a left and right hand side. On the left hand side, is the title "Generate images using Stable Diffusion" followed vertically by 3 single stacked bar charts (labelled B1-3)- the top one showcases gender, the middle one showcases race and the bottom one showcases age groups. Below, there is a button, which can be pressed to "generate baseline images"(labelled C), followed by an image gallery(labelled D), which depicts a row of 5 images containing representations of light skin tone, male software engineers. On the right hand side there is the title "Generate images based on your demographic representation", followed by a 4 worldview buttons "Parity", "US demogr.", "Absolute cat.", "Relative to the baseline" (labelled E). In the image the button "Parity is selected".  To the right of the worldview buttons is a slider titled "Proximity to Parity", which is greyed out (labelled F). Below there are 2 types of components that are alternated: a group of checkboxes where users can "Select what genders to represent" and can choose between "Male" and "Female" (labelled G1); a single stacked bar chart representing selected gender distributions which is half male and half female (labelled H1); a group of checkboxes where users can "Select what ethnicities to represent" and can choose between "Black", "East Asian", "Hispanic", "Indian", "Middle Eastern", "Southeast Asian" and "White" (labelled G2); a single stacked bar chart representing selected ethnicity distributions which are equally proportional amongst all ethnicities (labelled H2); a group of checkboxes where users can "Select what age groups to represent" and can choose between ``0--2'', ``3--9'', ``10--19'', ``20--29'', ``30--39'', ``40--49'', ``50--59'', ``60--69'', ``70+'' (labelled G3); a single stacked bar chart representing selected age group distributions which are equally proportional amongst all ages (labelled H3). Below that is  a button, which can be pressed to "generate edited images", followed by an image gallery(labelled L), which depicts a row of 5 images containing representations of diverse skin tone, gender and age software engineers.}
\label{fig: Dashboard}
\end{teaserfigure}

\maketitle

\section{Introduction}
A \textit{worldview} comprises a set of beliefs, values, and concepts used to interpret and make sense of the world~\citep{Dilthey1988}.
It is formed by gathering data from the world and learning a conceptual framework of relationships between the information.
As humans, we use our senses to perceive information and then filter and organize it to construct and update our worldview.
Given a setting, we use our worldviews to guide our decision-making process.
For example, choosing to dine at a local restaurant may be influenced by a person's positive past experiences and values of supporting small businesses in their community.
However, this process is not exclusive to humans; data-driven technologies, like generative text-to-image (TTI) models, also learn worldviews.
Models develop worldviews by learning connections between concepts from their training data that allow them to minimize an optimization function. 
Through this process, they learn to reflect a worldview that informs the space of outputs they can produce.

However, the worldviews reflected in TTI models do not always align with human expectations.
Common TTI models, like Stable Diffusion~\citep{rombach2022high} and DALL-E~\citep{ramesh2022hierarchical}, learn their worldview by training on data scraped from the internet.
Each of these dataset and modeling choices influences the worldview the model will reflect.
For instance, Stable Diffusion~\citep{rombach2022high} is trained on the image-text pairs~\citep{schuhmann2022laion} from a repository of web-crawled data~\citep{commoncrawl} that is filtered using CLIP~\citep{radford2021learning}.
Each of these data and modeling choices influences the worldview the model will reflect.
Stable Diffusion's data are captured from a predominantly Western male world~\cite{luccioni2021s, bender2021dangers} of younger English-speaking individuals from developed countries~\cite{worldbank}, contain stereotypical or under-specified information about minority cultures and groups~\citep{birhane2021multimodal, kay2015unequal, mandal2021dataset, otterbacher2017competent, naik2023social}, and are filtered using a classification model known to be more accurate in identifying white people~\citep{wolfe2022markedness}.
As a result, Stable Diffusion and other TTI models learn a worldview that reflects these choices and have been observed to reinforce and amplify biases and stereotypes in their generated images~\citep{bianchi2022easily, friedrich2023fair, naik2023social}.

To address the misalignment between TTI model and user worldviews, research has proposed strategies to modify TTI models, such as adding guardrails to remove not-safe-for-work (NSFW) concepts~\citep{dallemitig}, designing prompts and datasets that target ``fairer'' outcomes~\citep{bansal2022well}, and fine-tuning models to produce outputs representative of particular cultures~\citep{japanese_stable_diffusion, indigenous2023poc}. 
However, given the scale of the datasets and models, misalignment may remain even after these strategies are applied, and addressing the lack of representation for all user groups is a significant computational and resource challenge.
To this end, research efforts have aimed at developing post-processing techniques to steer TTI model outputs away from inappropriate concepts~\citep{Schramowski_2023_CVPR} and reduce representational biases~\citep{friedrich2023fair}. 
These methods enable users to shift the outputs of the diffusion process along chosen semantic directions to better align with their desired worldview; however, they have been primarily used in a limited capacity (e.g., enforcing gender parity for occupation bias) that does not reflect the diversity of worldviews a human might want represented in any given context.

To enable user control of the worldviews reflected in TTI-generated images, we develop \worldviewer.
\worldviewer is an interactive visual tool that exposes the worldview of existing TTI models and provides tools for users to align TTI outputs with their values and goals.
Leveraging tools to predict image demographics, \worldviewer exposes the gender, race, and age demographics of TTI-generated images.
If the demographics do not align with a user's worldview, \worldviewer provides controls to adjust the generated outputs to meet their expectations.
In this way, \worldviewer enables analysis of underlying TTI demographics and encourages users to make a conscious decision about which worldview to apply.
We demonstrate how \worldviewer encourages users to consider the worldview represented by TTI models via an evaluative user study and two case studies of common image generation tasks.
Through a think-aloud study with 18 users (\cref{sec: userstudy}) with diverse worldviews spanning varied job titles, unique cultures, and race, age, and gender categories, we find that \worldviewer shifts the process of worldview choice from under-the-hood decisions to a conscious end-user decision.
Furthermore, through case studies (\cref{sec:casestudy-section}) we demonstrate \worldviewer's applicability to common TTI tasks, including generating more inclusive representations and exploring alternate representations of community.
These results illustrate that \worldviewer represents meaningful progress towards user awareness and control over the values and identities output by AI models.

We contribute: 
\begin{itemize} 
    \item \textbf{A worldview formalism} that describes aspects of a worldview, how it influences image generation, and how existing editing tools modify worldviews;
    \item \textbf{Identified worldview system design opportunities} for tools to help users from diverse backgrounds identify and modify TTI model worldviews;
    \item \textbf{\worldviewer}, a tool to expose and expand the demographic worldview reflected in TTI-generated images; and,
    \item \textbf{Evaluative user studies and case studies} that demonstrate how \worldviewer scaffolds and accelerates real-world analysis of TTI worldviews and improves the representation of diverse worldviews.
\end{itemize}

\worldviewer is freely available as open-source software at \url{https://github.com/zoedesimone/DiffusionWorldViewer}. 
\newline

\noindent
\textit{Disclaimer}: This paper showcases various worldviews and stereotypes through images and employs gender and race classifiers that some readers may find offensive. We emphasize that our objective is to study and counter these within generative models. We neither seek to discriminate against any identity groups nor endorse the gender and race categorizations employed by these classifiers.

\section{Related work}

\subsection{Incorporating Worldviews into Technology}
Our research builds upon critiques challenging the notion of a singular, universal user representation~\citep{schlesinger2017intersectional}, which, as highlighted by \citet{schlesinger2017intersectional}, tends to homogenize individuals by neglecting differences and unique perspectives~\citep{dourish2012ubicomp, costanza2020design}.
To incorporate underrepresented worldviews and consider the intersectional identity of users, we formalize the concept of a \textit{worldview}.
This framework posits that users posses diverse worldviews shaped by their lived experiences~\citep{Dilthey1988}.
To support diverse worldviews in TTI model outputs, we design \worldviewer to give users explicit control of the race, age, and gender demographics in generated images.
We argue that \worldviewer represents meaningful progress towards user awareness and control over the values and identities output by AI models.

\subsubsection{Deciding whose worldviews are included in technology}
The evolution of Human-Computer Interaction (HCI), from its first to fourth wave, reflects a transformative shift in the conceptualization of the user.
After three value-neutral waves focused on personal computing, group work, and ubiquitous computing~\citep{harrison2007three}, the fourth wave has adopted an active stance with a ``primary focus on politics, values, and ethics''~\citep{ashby2019fourth}.
Within this paradigm, the challenge for HCI lies in determining user values and navigating their interactions with technology~\citep{sellen2009reflecting}.
There has been growing acknowledgement that cultural identity and tradition could be lost to technological standardization~\citep{shen2006towards} and the need to incorporate diverse worldviews in technology design by showcasing dissenting voices~\citep{gordon2022jury}, designing value-sensitive algorithms~\citep{scheuerman2020we}, community engagement~\citep{costanza2020design}, and design frameworks that embrace specific values~\citep{bardzell2010feminist, keyes2019human, ogbonnaya2020critical,scheuerman2020we, d2023data}.

Data collection, in particular, has been scrutinized for its potential to include or exclude particular communities or perspectives~\citep{d2023data, scheuerman2020we}.
While, in practice, datasets are considered impartial, researchers have found that datasets are often imbued with political values~\citep{scheuerman2020we} and decontextualize the cultural norms of information~\citep{denton2021genealogy}.
In response, calls for data feminism emphasize equitable collection, inclusion, and portrayal of diverse perspectives~\citep{d2023data}.

\subsubsection{Interfaces for exposing worldviews embedded in technologies}
The worldviews AI models learn from their datasets and training procedures are often hidden, so HCI and AI research has focused on interpreting model worldviews and visualizing them to end users.
Tools like Know Your Data~\citep{knowyourdata}, FairVis~\citep{cabrera2019fairvis}, the Embedding Comparator~\citep{boggust2022embedding}, and TCAV~\citep{kim2018interpretability} help users in understanding data quality, model predictions, and potential biases, while interactive blogs explore worldviews embedded in datasets~\citep{pairexplorables}.
Most related to \worldviewer is work incorporating user context in interpretability.
In particular, \citet{gordon2022jury} expose different perspectives that are present in human annotated datasets, allowing users to compare the distribution of annotator opinions.
The importance of context in ML evaluations is further explored in Kaleidoscope~\citep{suresh2023kaleidoscope} grounded in the notion that desired model behavior differs across contexts, and preference over ground-up model evaluation.

In the context of TTI worldviews, preliminary work has explored the design of interfaces to visualize underlying biased and societal representations of TTI models~\cite{luccioni2023stable, friedrich2023fair}. 
For example, Stable Bias~\citep{luccioni2023stable} allows users to compare outputs of various diffusion models across professions, compare facial features, and cluster by skin tone.
Relatedly, Fair Diffusion Explorer~\citep{friedrich2023fair} presents an interface where users compare the outputs of biased and gender-parity corrected images across professions.
In contract, \worldviewer's interface allows users to directly edit TTI model outputs according to their worldview and the needs of their task.

\subsection{Changing Worldviews in TTI Models} \label{sec: changing-worldviews}
Attempts to reduce bias and harm in TTI models span the model development process, from filtering datasets to remove harmful content~\citep{yang2020towards, nichol2021glide, schramowski2022machines, rombach2022high}, enforcing constraints during training~\citep{berg2022prompt}, removing concepts from the model after training~\citep{gandikota2023erasing, gandikota2023unified}, and post-processing model outputs at inference time~\citep{bansal2022well, brack2023sega, friedrich2023fair, Schramowski_2023_CVPR}.
In \worldviewer, we focus on post-processing techniques because they require less technical expertise and model retraining, which can pose a significant barrier for novice users.

\subsubsection{Pre-processing}
Model stereotypes and biases are closely tied to biases present in their training datasets~\citep{paullada2021data}.
Studies have identified biases such as Amerocentric and Eurocentric representation~\citep{shankar2017no} and discrepancies in prediction results for population subgroups~\citep{yang2020towards}.
In turn, calls for increased transparency, accountability, and geo-diversity in dataset development~\citep{denton2021genealogy, shankar2017no} have led to algorithmic interventions that promote demographic balance in training datasets~\citep{shankar2017no, merler2019diversity, yang2020towards, karkkainen2021fairface} and more demographically balances datasets and models~\citep{merler2019diversity, karkkainen2021fairface}.

\subsubsection{Training}
Model training decisions, such as the choice of loss function, shape the worldview reflected in ML models by determining the learned patterns from the data.
For instance, Stable Diffusion prioritizes image fidelity, quality, and realism in its loss functions, reinforcing patterns and stereotypes from the training data~\citep{rombach2022high}.
Efforts to shift worldviews during training time have focused on including fairness terms in model objective functions~\cite{martinez2020minimax} and dataset sampling techniques to improve representation~\cite{mandal2020ensuring, roh2020fairbatch, yan2022forml}. 

\subsubsection{Fine-tuning}
Since large-scale datasets are critical to the success of diffusion models and collecting data is challenging and expensive, some worldview editing approaches have focused on fine-tuning.
In particular, \citet{gandikota2023erasing, gandikota2023unified} remove NSFW concepts from diffusion models via fine-tuning, and online TTI model communities have used fine-tuning to create culture-specific Stable Diffusion variants~\cite{japanese_stable_diffusion, indigenous2023poc}.

\subsubsection{Post-processing}
Post-processing techniques focus on editing the model's output without changing the model.
Popular post-processing techniques in TTIs include prompt engineering~\citep{liu2022design} and semantic guidance~\citep{brack2023sega}.
Prompt engineering modifies the model's prompt to target a specific outcome (e.g., images with a particular demographic); however, it can require extensive trial and error to generate a prompt that produces the correct output.
On the other hand, semantic guidance guides the diffusion process towards user-specified concepts during image generation~\citep{brack2023sega}.
Semantic guidance has been used to successfully suppress inappropriate NSFW image regions~\citep{Schramowski_2023_CVPR} and shift outputs towards gender parity~\citep{friedrich2023fair}.
\worldviewer builds on semantic guidance techniques because it allows users to reflect their worldviews in TTI models without additional training and regardless of the limits of the model's worldview.

\section{Worldview Formalism}
\label{sec:worldview-formalism}

The idea of a worldview was popularized by \citet{Dilthey1988} who introduced the term to theorize how humans understand things.
Under \citeauthor{Dilthey1988}'s framework, a worldview is a comprehensive set of beliefs, values, and concepts that an individual or group uses to interpret and understand the world around them.
A person's worldview guides the collection and interpretation of information captured from the world and subsequently influences how they act.
Since worldviews are shaped by our historical and cultural context and personal life experience, they can vary greatly between individuals.
Highlighting differences between people's worldviews can reveal the subjectivity of human interpretations, provide unity, and help us make sense of the diverse experiences we encounter~\cite{Dilthey1988}.

Following \citet{pairexplorables} and \citet{gordon2022jury}, we extend \citeauthor{Dilthey1988}'s worldview framework to include AI agents.
Like humans, AI agents use their internal parameters to interpret information from the world and take actions that achieve their goals.
An agent's worldview is shaped by its ``experience'' of the world, including the data it has seen, its training procedure, and its optimization function.
As a result, agents trained on different datasets or with different objectives may learn different worldviews.
For instance, TTI models often learn a Westernized worldview given their training data primarily contains English text and white, male, and Western images.
However, TTI models finetuned on Japanese data have been shown to learn a worldview that better reflects Japanese culture~\citep{japanese_stable_diffusion}.
In the same way that understanding another person's worldview can help us make sense of their actions, understanding an agent's worldview helps us calibrate our expectations of their capabilities and interact with them effectively.

We formalize our notion of a worldview to describe how a TTI agent forms a worldview and the mechanisms we have to modify them.
We take a worldview to contain two essential components, \capta\,---\, the information provided to the agent (e.g., sensory inputs for humans or (\texttt{text}, \texttt{image}) pairs for TTI models)\,---\, and \arrows\,---\, the learned relationships between the \capta.
In this framework, a worldview is a learned relationship between \capta and \arrows.
\[ \texttt{learn}(\capta) \rightarrow \arrows \]
\[ \worldview: \{\capta, \arrows\} \]

\subsection{\textcolor{CaptaColor}{CAPTA}}
\capta are the information provided to an agent.
Instead of the term \textit{data}, which is often viewed as passively `given' and waiting to be collected, our work builds on the definition of \capta which is understood as being actively `taken' or `captured'.
This implies a more participatory role in the creation of information, where the observer or agent plays a central role in determining what is significant and worth capturing~\cite{Drucker2011HumanitiesAT, kitchen21}.
As such, \capta are an inherently partial representation of the world, influenced by our perspectives, biases, and the specific context of our inquiry.
For humans, \capta are experiences. Each person constructs a unique understanding of the world by selectively capturing and processing information based on personal biases, perspectives, and context.
For TTI models, \capta are (\texttt{image}, \texttt{text}) pairs.
The process of creating a TTI dataset involves actively selecting a subset of text and images, and this process is influenced by the model designers' perspectives, the intended use of the model, and the cultural and contextual factors at play.

\subsection{\textcolor{ArrowColor}{ARROWS}}
\arrows symbolize the dynamic connections between \capta that represent how an individual interprets, relates, and gives meaning to their experiences.
These \arrows form an `understanding' or a set of beliefs about the world that is unique to each individual and shaped by their subjective experiences and interpretations.
\[ \text{\arrow}: \capta \rightarrow \capta \]
For humans, \arrows are learned associations that give rise to actions.
Actions are the result of prior experiences (\capta) that have formed a person's understanding of the world (\arrows).
TTI models are a collection of \arrows, symbolizing the learned mappings between textual inputs and visual outputs.
These mappings develop dynamically during the training process, where a TTI model absorbing patterns, correlations, and associations from the specific dataset it is exposed to. 
These differences lead to a unique worldview with interpretative strengths and biases, with some models attuned to specific visual styles or textual nuances.

\subsection{Using and Estimating Worldviews}
Once a worldview is established, agents (i.e., humans or models) use their worldviews to take action (e.g., generate an image) under a given \context.
A \context does not change the agent's worldview, but it impacts the output an agent takes.
For instance, a person may behave more formally at a professional gathering than at dinner with friends.
The setting has not changed the person's worldview, but it has operated on their worldview to elucidate different actions.
Similarly, for TTI models, the text prompt provides \context that guides the model to a particular output.
Different prompts do not update the model's internal representations or worldview, but they allow us to access different model behaviors.
However, the model's worldview still plays an important role, as it dictates the space of possible actions.
For instance, a person that does not speak Spanish can not have a fluent conversation even if they were surrounded by Spanish speakers, and a TTI model can not output high-quality images from prompts that are substantially outside of their training distribution.
Together the \context and the worldview define the action of the agent.
\[ \generate(\worldview, \context) \rightarrow \texttt{output}\]

Ideally, we would like to extract the agent's worldview to understand how it differs from our worldview and estimate it's behavior so we can work together most effectively.
However, worldviews are abstract concepts that are hard to communicate, so, instead, we estimate worldviews by observing an agent's behavior in \context and comparing it to our expectations.
\[ \estimate  = \generate(\textsc{agent\_worldview}, \context) \widesim{} \]
\[ \generate(\textsc{user\_worldview}, \context)\]
This occurs regularly between humans as they interact with others and observe their behavior.
In TTI models, estimation can occur overtime as users build intuition by prompting the model and serendipitously identifying areas where the generated images align or conflict with their expectations.
Or, it can result from directly testing the model with a challening prompt to estimate a particular aspect of it worldview.

\subsection{Operations on Worldviews}
\label{sec:worldview-operations}
Often we want TTI models to \generate a specific output that aligns with our goals and worldviews.
Participants in our user study (\cref{sec: userstudy}) often had a specific image in mind that they expected the model to output.
Since a TTI model's output is a function of its \worldview and the \context we provide, existing techniques modify TTI model outputs by adjusting its \worldview (i.e., by changing its \capta, \arrows, or both) or \context.
By describing the components of a worldview, our formalism allows us to describe worldview modifications as functions on \capta, \arrows, and \context.
It allows us to formalize existing algorithms (\cref{sec: changing-worldviews}) and provides scaffolding to generate new algorithms.

\subsubsection{Pre-processing}
The pre-processing phase of TTI model development focuses on generating \capta that will train or finetune the model.
In this phase, adjustments to the training set can be made to diversify or modify the \capta. 
For instance, if the original training data primarily represents a specific demographic or geographic area, but the model is intended for broader use, this stage offers the opportunity to introduce new \capta that encompasses a wider range of contexts. 
By integrating data from diverse demographics and locations, the model's output can become more representative of the varied contexts it is designed to serve. 

\subsubsection{Training}
During the training process, model developers select the training parameters, such as the objective function, model architecture, and data sampling procedures.
These choices modify the \arrows the model will learn between the \capta. 
For instance, in TTI models, the objective function typically aims to balance the fidelity of the text descriptions and the quality and realism of the generated images. 
Altering the objective to give more weight to certain types of \capta can lead to models that learn different \arrows and potentially embody different worldviews.

\subsubsection{Post-processing}
After a model is trained, it is much harder to alter its worldview since the \capta that guided its development have been set.
However, there are still techniques available that can tweak this mapping without the need for additional training.
Some existing techniques, like inpainting~\cite{zeng2020high} and prompt engineering~\cite{liu2022design}, modify the \context given to the model, while others, like semantic guidance~\citep{brack2023sega}, temporarily upweight particular \arrows in the model to increase the likelihood of producing a particular output.
Generally, post-processing approaches are advantageous due to their lower resource requirements and provide a degree of flexibility in influencing how the model generates outputs.

\paragraph{Prompt engineering}
In prompt engineering, users modify the text prompt to obtain an output image that aligns with their worldview.
Prompt engineering modifies the model's \context, which activates a different aspect of a model's worldview, resulting an image that better reflects the users expectation. 
Since the model's output is a function of its \worldview and \context, prompt engineering often requires an iterative trial and error process to find prompts that, with the model's worldview, achieve the user's desired output and build intuition for future prompts.
\[ \texttt{prompt\_engineer} = {\generate(\worldview, \textcolor{ContextColor}{\texttt{prompt}_i})}\]
\[ \, \forall \, i \text{ until } \estimate(\textcolor{ContextColor}{\texttt{prompt}_i}, \worldview, \textsc{user\_worldview}) \]

\paragraph{Inpainting}
Inpainting generates outputs by applying different prompts to specific image regions.
Users create semantic masks that confine text prompts to alter designated areas of the image.
This process blends elements from multiple prompt \textcolor{ContextColor}{\textsc{contexts}}, each of which trigger a different aspect of the TTI model's worldview.
The resulting composite image reflects a tailored worldview more closely aligned with the user's specific intentions.
\[ \texttt{inpaint} = \sum_{i}{\texttt{generate}(\worldview, \textcolor{ContextColor}{(\texttt{prompt}_i, \texttt{region}_i}))} \]

\paragraph{Semantic guidance}
Semantic guidance allows users to steer image generation towards \concept~\citep{brack2023sega}.
In this form of editing, users specify a \concept (e.g., \textit{female}) and semantic guidance identifies the \concept's \arrow in the model's latent space and shifts the model's output towards the \concept \arrow, enabling the integration of target \concepts into generated images.
By manipulating these \arrows during image generation, semantic guidance allows for targeted adjustments to the generated images that align more closely with the user's intended concept.

\[ {\arrows'} = \texttt{semantic\_guidance}(\concepts) \]
\[ \worldview' = \{\capta, \arrows'\}\]
\[ \texttt{generate}(\worldview', \textcolor{ContextColor}{\texttt{prompt}}) \]

\section{\worldviewer Challenges and Design Opportunities}\label{sec:goals-section}
To understand how to best support worldview analysis and editing, we explored users' needs and challenges through informal discussion with TTI users and ML researchers and identified design opportunities where worldview tools could provide value to TTI users.

\begin{enumerate}[label=\textbf{D{{\arabic*}}}]
\setlength\itemsep{0.3em}
\item  \textbf{Reveal TTI models' worldviews to users.}\label{design-goal:reveal-worldviews}\\
Since TTI models use their \worldview and prompts (\context) to \generate images, users can \estimate aspects of a model's worldviews by observing its outputs.
However, inferring a complete understanding of a model's worldview from individual prompts can be challenging and time-consuming.
Many users do not actively consider the model's underlying worldview unless the generated images clash with their own perspectives. 
Even when conflicts arise (e.g., bias), users often attribute them to one-off mistakes, assuming the model's worldview aligns with theirs~\citep{gray2011worldviews}.
Our conversations revealed that gaining a comprehensive understanding of a model's worldview was rare, with expert users aware of limited conflicts (e.g., race and gender biases) and novice users oblivious to any worldview conflicts
To address this, worldview systems should transparently surface models' worldviews, so users of all experience levels can consciously reflect on and challenge the worldviews in TTI models. 

\item \textbf{Modify TTI models to reflect users' diverse and individual worldviews.}\label{design-goal:reflect-users}\\
TTI model users have diverse worldviews they want to see expressed in TTI models.
However, as users \generate images, they often find the outputs do not align with their expectations (\estimate).
A common issue arises from demographic biases within TTI model worldviews, reflecting problematic \arrows between \capta of underrepresented groups and harmful stereotypes~\citep{bianchi2022easily, friedrich2023fair, naik2023social}.
Similarly TTI worldviews are often constrained to \capta from particular geographies and cultures, while users \textit{``want a model to understand [their] culture, identity, and unique expressions, such as slang''}~\citep{japanese_stable_diffusion}.
Although there have been efforts to increase gender representation~\citep{friedrich2023fair} and fine-tune TTI models for specific cultures~\citep{japanese_stable_diffusion, indigenous2023poc}, these solutions do not reflect the rich diversity of existing worldviews. 
Worldview systems should expand editing controls to better represent and reflect the multitude of worldviews in the TTI community. 

\item \textbf{Support TTI editing for novice and expert users.}\label{design-goal:editing}\\
When TTI models' and users' worldviews do not align, users turn to editing techniques such as prompt engineering~\citep{liu2022design}, inpainting~\citep{rombach2022high}, and semantic guidance~\citep{brack2023sega} (\cref{sec:worldview-operations}) to guide the outputs toward their expectations
While these techniques can successfully shift model outputs, they often require manual user effort, skillful tool usage, and have limited changes (e.g., gender parity~\citep{friedrich2023fair} or NSFW images~\citep{Schramowski_2023_CVPR}).
However, TTI model users have diverse experiences and use cases, ranging from daily professional tasks to occasional recreational use.
Novice users prefer tools that work seamlessly without prior TTI model skills, while expert users expect customization and integration into their complex workflows. 
TTI model worldview analysis and editing tools should be accessible to users of all experience levels, striking a balance between code-free usability and code-based customization for integration into existing TTI frameworks.

\item \textbf{Prevent adversarial and harmful worldview editing.} \label{design-goal:harms}\\
TTI model have been shown to express harmful concepts and stereotypes~\citep{bianchi2022easily}.
While harm mitigation efforts have developed guardrails~\citep{rombach2022high, Schramowski_2023_CVPR} that prevent concepts from appearing in models, models can still generate stereotypical and harmful representations~\cite{bianchi2022easily}.
In our conversations, practitioners note issues like female sexualization in TTI models and emphasize the need for limitations on such problematic representations.
While worldview systems aim to increase control over the worldviews reflected in model outputs by allowing users to modify generated images, the challenge lies in striking a balance between enabling user freedom of expression and reducing the risk of generating harmful worldviews.

\end{enumerate}

\section{Design of \worldviewer}
\worldviewer allows users to analyze and edit the underlying worldviews in TTI models based on their personal viewpoint and values.
Critically, it is not limited to specific text prompts or to a single definition of a worldview, and is capable of supporting a wide range of text prompts and worldviews.
\worldviewer is designed in two parts: the front-end, where we define the organization and design of the dashboard, and the back-end, where we define the worldview analysis and editing methodologies.

\subsection{Back-end Design}
To address design goal~\ref{design-goal:reveal-worldviews}, \worldviewer facilitates the \estimate of the worldview reflected by the TTI model by classifying the demographics of Stable Diffusion model output. \worldviewer allows users to increase representation of alternative worldviews in images without burdening them with prompt engineering, or other manual techniques (Design Goal~\ref{design-goal:reflect-users}).

\subsubsection {Surfacing Underlying Worldviews}
Given that TTI outputs are not labelled, analyzing worldviews in TTI requires quantifying these concepts in image outputs through an \estimate. However, translating these abstract concepts into quantifiable and actionable elements to analyze and edit requires the ability to classify and categorize, which is inevitably reductive to some extent. 

To allow users to query the worldview of the model, we utilize classifiers and report summary distributions over a subset of five model image outputs for every input text prompt (\context). In keeping with prior literature showcasing the stereotypes in Stable Diffusion~\citep{bianchi2022easily, friedrich2023fair, naik2023social}, we focus on a subset of categories that shape a worldview, namely demographic characteristics\,---\,gender, race and age. While there are other categories that are important in shaping our worldview, such as background, geography and culture, and we would ideally include all of these categories in \worldviewer, existing image classifiers limit the categories we can discriminate.

We compute demographics using the FairFace classifier~\citep{karkkainen2021fairface}, which predicts the likelihood of an image belonging a certain gender, ethnicity and age. Categories include 7 race groups (\texttt{white}, \texttt{black}, \texttt{indian}, \texttt{east asian}, \texttt{southeast asian}, \texttt{middle eastern}, and \texttt{latino}), 9 age groups (\texttt{0--2}, \texttt{3--9}, \texttt{10--19}, \texttt{20--29}, \texttt{30--39}, \texttt{40--49}, \texttt{50--59}, \texttt{60--69}, \texttt{70+}), and 2 gender groups (\texttt{male} and \texttt{female}). 
Top-class predictions of race, gender and age for each image are grouped and used to analyze the underlying worldview of a given text prompt.

\subsubsection {Editing Toward User-Specified Worldviews} \label{sec:backendediting} %
Informed by Design Goals~\ref{design-goal:reflect-users}~and~\ref{design-goal:editing}, \worldviewer allows users to edit model worldviews according to their personal values and viewpoints.
To allow real-time editing, \worldviewer relies on post-processing editing techniques that modifying existing \arrow mappings amongst \capta using an already trained model.
While we initially explored several post processing operations on worldviews, including prompt engineering and inpainting, we use \(\texttt{semantic\_guidance}\) because it allows for fine-grained changes in outputs while preserving the original composition of the TTI image.

To guide users in worldview editing, we provide two example demographic distributions\,---\,\textit{demographic parity} and \textit{US demographics}.
Demographic parity allows users to impose parity across the FairFace gender, race and age groups, while US demographics enforces FairFace gender, age, race categories based on the percentage demographics of the US population from the Census~\citep{Census}.
Beyond these present, users can perform \textit{absolute editing} where they explicitly specify race, age, and gender distributions that determine which edits to apply to each image.
Alliteratively, \textit{relative editing} allows users to specify how much to modify Stable Diffusion's baseline demographics towards parity.
The baseline Stable Diffusion demographic distributions are computed as described in \cref{sec:backendediting}.

For each of these editing techniques, the demographic distributions are used to determine weighted dice throws to stochastically generate natural language \concepts to edit each image. 
A single keyword is generated for each of the gender, age and race categories to form an editing triple such as (``\textit{female person}'', ``\textit{black person}'', ``\textit{0--3 year old person}'') which is applied to modify the demographics of an image. The \concepts in the triple are applied as additive \concepts to the diffusion process though SEGA~\citep{brack2023sega}, which is steered in latent space towards the applied \concepts. 

Informed by~\ref{design-goal:harms}, our goal was to expose worldview controls to end users and empower them to generate more diverse representations and worldviews, while being mindful that our controls could be used adversarially. The challenge in exposing worldview controls lied in striking the balance between allowing enough freedom and control for users to express their views, whilst reducing the risk for harmful worldview generation. To that end, we designed relative editing to range from Stable Diffusion's baseline worldview to parity in its outputs. However, we thought giving users the flexibility of the absolute category would allow them to have more control over outputs. We further discuss this in \cref{sec:52}.

\subsection{Front-End Design}
Informed by the design constraints described in Goal~\ref{design-goal:reveal-worldviews}~and~\ref{design-goal:editing}, we develop \worldviewer, an interactive web application that leverages classifiers and demographic distribution plots to help users \estimate model worldviews, visualize the distribution of outputs, and test any text prompt (\context) to explore baseline model worldviews and apply different worldviews using \(\texttt{semantic\_guidance}\). %

\subsubsection{Interfacing with technical and non-technical users}
Creating a tool for power users and novices alike (Design Goal~\ref{design-goal:editing}) required choosing a solution that provided both a code free web-based tool as well as an accessible code base for power users to customize and integrate into their existing workflows. 
We initially explored a few key options, including A) {\em deployed web apps}, which require users to download the code and locally deploy the website, and therefore were a barrier for novice users, often compensated by creating demo websites with preloaded data (eg. Embedding Comparator \cite{boggust2022embedding}); B) {\em Python Notebooks} such as Colab, which require users to know how to code in Python and therefore is not adequate for novice users; C) {\em Symphony}~ \cite{composing-interactive-interfaces}, a framework that connects computational notebooks and web dashboards is a promising alternative to satisfy technical and non-technical users alike. However, the code for this tool is not publicly available. D) {\em Gradio library} \cite{abid2019gradio}, a Python library to host ML model interfaces online, popular among the TTI and Stable Diffusion community which can be hosted through Colab notebooks and publicly deployed, used in the {\em Stable Diffusion Web UI} \cite{sdwebui}. 
We settled on {\em Gradio} as it best supported \ref{design-goal:editing}. 
The web app can be hosted publicly through Hugging Face and can be privately launched via a Colab notebook, providing both code-free and code-based access points for the two user groups, as well as an opportunity to integrate the tool into existing Stable Diffusion web UIs, also built using Gradio. 

\subsubsection{Visually linking baseline and edited model outputs}
To showcase the underlying worldviews of Diffusion models and the tools to modify the worldviews of these models (Design Goal~\ref{design-goal:reveal-worldviews} and~\ref{design-goal:reflect-users}), \worldviewer introduces a side-by-side dashboard layout, which graphically displays the relationships between the baseline and modified model. The left side showcases baseline image outputs and demographics, while the right side showcases the editing pipeline, including worldview choices consisting of demographic distributions to impose onto the model as well as corresponding model outputs. This way the user can compare the \estimate worldview reflected by the original model outputs and the \estimate worldview reflected by the edit model outputs.

While we initially explored several UI designs and organizations such as placing editing controls in side bars, several panels to visualize the distributions, and condensing images outputs and classifier information horizontally, we decided against these organizations as they did not facilitate back-and forth between the two model outputs. We opted instead for a side-by side dashboard layout, inspired by prior stable diffusion dashboards such as FairDiffusion \cite{friedrich2023fair}. This design allowed for a visual comparison of original and edited outputs, as well as a visual parallel between the classifiers used to help \estimate the underlying worldview reflected by the model  (Figure \ref{fig: Dashboard}D and L) and the controls to steer \(\texttt{semantic\_guidance}\) (Figure \ref{fig: Dashboard}B and H). Furthermore, while our initial prototypes included a text prompt input for each of the models, we simplified and shared components across the two sides when possible to encourage back-and-forth exploration across the two models.

The baseline view (left) allows users to explore the worldviews reflected in the Stable Diffusion model. After entering a text prompt (\context) (Figure \ref{fig: Dashboard}A) users can see demographic distributions for gender, race and age visualized in 3 stacked bar charts (Figure \ref{fig: Dashboard}B), as well as the output images which are loaded into an image gallery upon generation (Figure \ref{fig: Dashboard}D). 

The editing view (right hand side) allows users to explore and visualize the impact of different editing techniques on the model's reflected worldview. At the top of the dashboard a user can specify one of 4 worldview editing options to apply to the model: parity, US demographics, absolute and relative editing (Figure \ref{fig: Dashboard}E). Below, the user can visualize the demographic distributions of the specified worldview edit, through 3 stacked bar charts depicting the distributions for gender, race and age representing the demographics of the selected worldview (Figure \ref{fig: Dashboard}H). Above each bar chart is a series of check boxes, which can be selected if a user wishes to explicitly specify the gender, age and race groups through the absolute editing technique (Figure \ref{fig: Dashboard}G). Below the visualizations is an image gallery which loads edited images as they are generated (Figure \ref{fig: Dashboard}L). In the case a user selects the relative editing technique, a slider is enabled which allows users to specify how much they want to modify demographic distributions towards parity (Figure \ref{fig: Dashboard}F). 

\begin{figure*}
   \includegraphics[width=\linewidth]{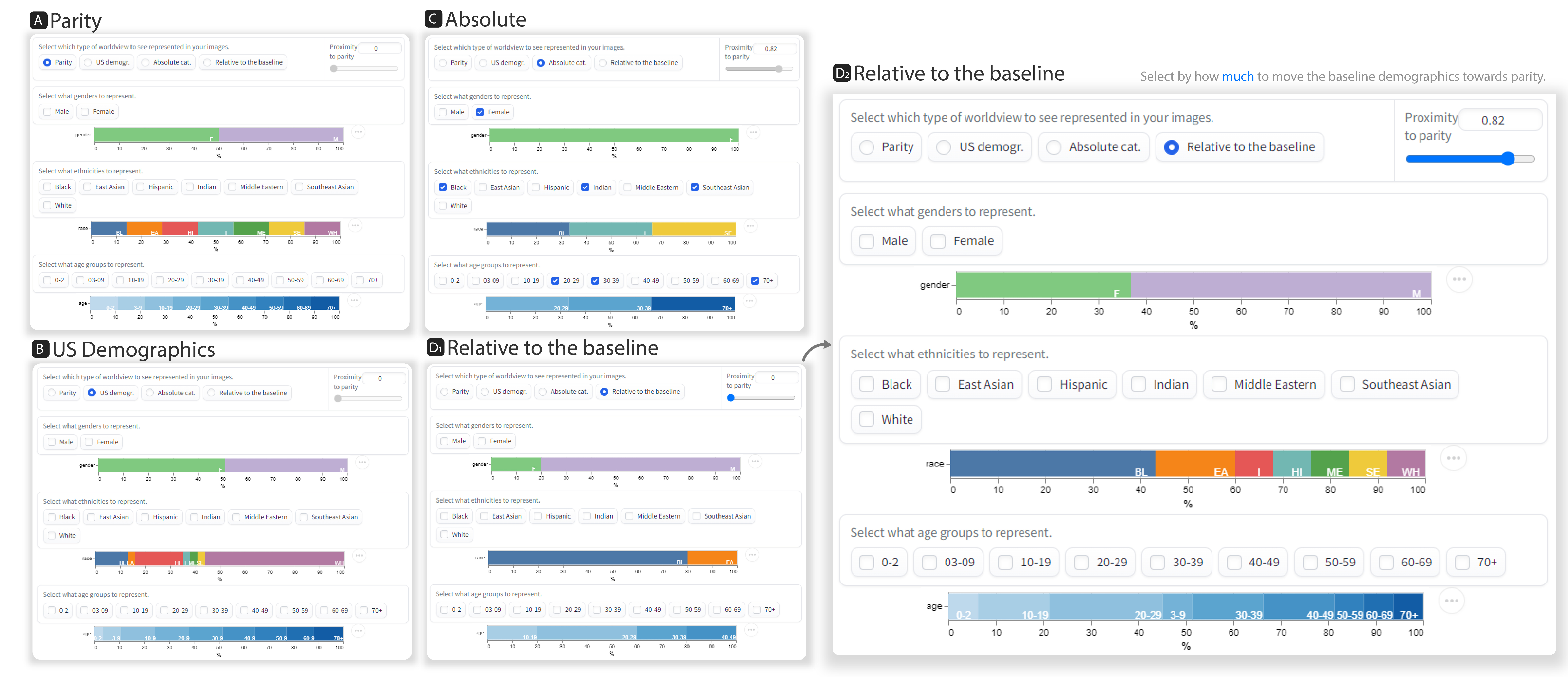}
   \caption{The \worldviewer allows users to specify a new worldview to impose onto the model outputs: Parity (A), US demographics (B), Absolute (C) and Relative (D1-2) editing techniques. Using the Relative editing technique users can specify given the baseline demographic distributions from the baseline SD model, by how much they want to modify the distributions towards parity.}
   \Description{Screenshots of the \worldviewer 4 editing methods and a zoom in of the "Relative to the baseline" editing methods. Each editing dashboard screenshot includes 4 worldview buttons "Parity", "US demogr.", "Absolute cat.", "Relative to the baseline". To the right of the worldview buttons is a slider titled "Proximity to Parity". Below there are 2 types of components that are alternated: a group of check boxes where users can "Select what genders to represent" and can choose between "Male" and "Female" ; a single stacked bar chart representing selected gender distributions; a group of check boxes where users can "Select what ethnicities to represent" and can choose between "Black", "East Asian", "Hispanic", "Indian", "Middle Eastern", "Southeast Asian" and "White"; a single stacked bar chart representing selected ethnicity distributions; a group of check boxes where users can "Select what age groups to represent" and can choose between "0-2", "3-9", "10-19", "20-29", "30-39", "40-49", "50-59", "60-69", "70+"; a single stacked bar chart representing selected age group distributions. 
   The five screenshots, in order depict the following:

   In the first image (A) the button "Parity is selected", and the "Relative to the baseline" slider is greyed out. Stacked bar charts depict gender, ethnicities and age groups categories to be same in size. No demographic check boxes are selected.
   
   In the second image (B) the button "US demogr" is selected, and the "Relative to the baseline" slider is greyed out. Stacked bar charts depict gender, ethinicities and age groups cateogires reflect the US population demographics relative to age, ethnicity categories. No demographic check boxes are selected. 
   
   In the third image (C) the button "Absolute cat." is selected, and the "Relative to the baseline" slider is greyed out. Across demographic check boxes, the user has selected "Female", "Black", "Indian", "Southeast Asian", "20-29", "30-39", "70+". Stacked bar charts reflect checkbox selections depicting 100\% females, 1\\3 Black, Indian and Southeast Asian, and 1\\3 of the bar for each selected age group.
   In the fourth image (D1) the button "Relative to the baseline" category is selected, and the "Relative to the baseline" slider is enabled and set to zero. No demographic check boxes are selected. Stacked bar charts reflect the baseline distributions, depicting 1/4 females, 3/4 males; 3/4 Blacks and 1/4 East Asians, 1/2 20-29, 1/4 10-19, 1/4 30-39, 1/4 40-49. 
   
   In the fifth image (D2) the button "Relative to the baseline" category is selected, and the "Relative to the baseline" slider is enabled and set to 0.82. No demographic check boxes are selected. Stacked bar charts reflect the baseline distributions, depicting categories in close proportion to parity: 35\% females, 65\% males; 40\% Blacks, 15\% East Asians, and the remaining categories in equal proportion; age categories in closer proportion to parity.}
  \label{fig: editingoptions}
\end{figure*}

\subsubsection{Reducing the burden of representation in TTI}
Informed by Design Goal~\ref{design-goal:reflect-users} and~\ref{design-goal:editing}, our aim was to reduce users' burden of prompt engineering diverse representations by giving control over worldviews reflected by the diffusion model. To accomplish this goal we created a series of buttons where users could specify high-level what editing technique to modify the worldview reflected by the model (\ref{fig: editingoptions}, c), as well as more detailed options (\ref{fig: editingoptions}, g) where users could specify which genders, races and ethnicities to impose onto the model.

Two options \em{parity} and \em{U.S. demographics} load a pre-defined worldview to apply onto the model, and demographic distributions can be visualized via the stacked bar charts (\ref{fig: editingoptions}). The absolute worldview allows users to specify their worldview directly, via the selection of multiple categories amongst the options (\ref{fig: editingoptions}C). The relative worldview, allows users to specify by how much to modify baseline demographics, classified in the baseline, towards parity via a slider (Figure \ref{fig: editingoptions}, D1-2).

\section{Evaluation: User Study} \label{sec: userstudy} 
To evaluate \worldviewer, we conducted a user study with eighteen TTI users. Participants included designers, scientists, artists, programmers, and machine learning experts from academia and industry, representing 10 countries, and with a wide range of prior TTI experience (see \cref{tab:interview-participants}).
The goal of the study was to understand how practitioners would use \worldviewer to perceive worldviews embedded in existing TTI models and shift model outputs to reflect their worldviews.
We were also interested in understanding how exploring different worldview options provided by \worldviewer could help practitioners \estimate underlying worldviews in the models and increase their awareness of patterns and stereotypes in these.

\subsection{Study Design and Setup}
Each interview lasted between 45 minutes to 1 hour, and was structured in 3 sections. First, users were asked about their background, their TTI workflows, their tasks, the importance of different viewpoints and representation in their tasks, and any bias mitigation strategies they employ. 
Next, we sent the participants a one-time link that allowed them to access the \worldviewer web application in their own browser, and asked them to share their screen while they explored the tool. 
Users were shown the left-hand panel of the \worldviewer with baseline Stable Diffusion and were asked to test 2 to 3 prompts to \(\displaystyle \estimate \) the worldview of the model and determine in which ways it differed from their worldview. 
They were asked to think aloud and describe why they had chosen each prompt (\context), as well as their expectations of the images prior to generation as well as their reactions after images were generated. 
For each prompt we recorded the user's worldview \(\displaystyle \generate(\textsc{user\_worldview}, \context)\) (what they wished to see), their \(\displaystyle \estimate \) of the agent's worldview (what they actually saw as an output), as well as their expectation of the agent's worldview prior to generating any outputs (what they expected to see).

We then asked the participants to navigate to the Editor View, which showed the left and right-hand panel of the \worldviewer side by side. We introduced users to the 4 different worldview \( \texttt{semantic\_edit} \) techniques and asked them to select the editing technique that best represented their worldview. %
Users were then asked to explain their choice and prompted to apply the \( \texttt{semantic\_edit} \) and to comment on the images generated by the model. For each prompt we recorded the user's reaction to the \(\displaystyle \estimate \) worldview of the baseline model, their choice in editing guidance, as well as their reaction to the edited worldview \(\displaystyle \estimate \) after applying the changes. Next, participants were questioned about the potential advantages and drawbacks for each, as well any notions or categories missing from the options. Finally, participants were asked whether having access to \worldviewer's controls changed the way they thought about the worldview reflected by the model and whether they might integrate these tools in their own tasks.

The study was approved by our internal IRB. We recorded all interviews, with each participant's consent and participants were compensated with \$30 amazon gift cards. 
We recruited participants from an open call on social media, from an internal mailing lists at our institution and recruited well-known TTI professionals, choosing participants with different backgrounds spanning across geographies, industry/fields, demographics, and based on a range of prior use and expertise with TTI models (Table \ref{tab:interview-participants}).

\begin{table*}[t]
\centering
\caption{We evaluate \worldviewer with 18 TTI users. Each participant has a unique worldview shaped by their demographics, roles, and experience that inform their diverse TTI use cases. Participants self reported their demographics, roles, and usage ( --- indicates a participant preferred not to answer).}
\Description{A table that lists the metadata of the qualitative interview. Twelve participants are students and one is a researcher. Five participants work in art, seven in machine learning, two in design, one in robotics, and one in consulting. Three participants use TTI daily, nine use TTI occasionally, and one had never used TTI before.}
\label{tab:interview-participants}
\resizebox{\linewidth}{!}{
\begin{tabular}{l rlll ll ll}
 \textbf{ID} & \multicolumn{4}{l}{\textbf{Demographics}} & \multicolumn{2}{l}{\textbf{Role}} & \multicolumn{2}{l}{\textbf{TTI Experience}} \\
 & Age & Race & Gender & Country & Field & Title & Usage & Use Cases \\
\cmidrule(lr{0.1em}){1-1} \cmidrule(lr{0.1em}){2-5} \cmidrule(lr{0.1em}){6-7} \cmidrule(lr{0.1em}){8-9}
P1  & 18--29 & White          & Female & USA      & Art \& Architecture         & Student                  & Occasionally & Product design; event advertisement\\
P2  & 18--29 & Indian         & Male   & USA      & Machine Learning            & Student                  & Occasionally & Poster design\\
P3  & 18--29 & Asian          & Male   & China    & Machine Learning            & Student                  & Occasionally & Content creation; Illustration\\
P4  & 18--29 & White          & ---    & USA      & Machine Learning \& Art     & Student                  & Daily        &  Design research and ideation\\
P5  & 30--39 & ---            & ---    & Pakistan & Art \& Consulting           & Artist \& Researcher     & Daily        & Art; tattoo design; illustration \\
P6  & 18--29 & White          & Male   & USA      & Robotics                    & Student                  & Occasionally &  Art; Generating avatars for videogames\\
P7  & 30--39 & Black          & Male   & Kenya    & Business \& Design          & Student                  & Occasionally & Art; Design research;\\
P8  & 18--29 & White          & Male   & USA      & Machine Learning \& Physics & Student                  & Occasionally & Marketing; content creation  \\
P9  & 18--29 & Indian         & Female & India    & Design                      & Student                  & Occasionally & Digital art therapy\\
P10 & 18--29 & Asian          & Female & China    & Machine Learning            & Student                  & Occasionally & Art \\
P11 & 18--29 & Indian         & Female & India    & Art \& Architecture         & Student                  & Daily        & Architecture design renders \\
P12 & 18--29 & Indian         & Male   & India    & Machine Learning            & Student                  & Never        & ---\\
P13 & 30--39 & Latinx         & Male   & Chile    & Machine Learning            & Student                  & Occasionally & Illustration; Logo design\\
P14 & 40--49 & Middle Eastern & Male   & Egypt    & Art \& Design               & Designer                 & Daily        & Product design; architectural rendering\\
P15 & 30--39 & White          & Female & Greece   & Technology                  & Product Manager          & Occasionally & Design concept ideation; footwear design\\
P16 & 18--29 & White          & Female & Poland   & Art \& Design               & Architect \& Illustrator & Occasionally & Illustration\\
P17 & 40--49 & Middle Eastern & Male   & Egypt    & Design \& Machine Learning  & Designer                 & Occasionally & Architectural rendering \\
P18 & 30--39 & White          & Female & USA      & Machine Learning            & Software Developer       & Occasionally &  Synthetic image datasets\\
\end{tabular}
}
\end{table*}

\subsection{Qualitative Results}
We reviewed recordings of each study session and documented the prompt (\context), \(\displaystyle \estimate\) worldview, and \( \texttt{semantic\_edit} \). We analyzed each prompting instance, based on the user's {\em expertise}, their {\em exploration} of the baseline model \(\displaystyle \estimate\) worldview, their {\em motivation} for the  \( \texttt{semantic\_edit} \), and their {\em analysis} of the edited model \(\displaystyle \estimate\) worldview.
We define {\em expertise} as the user creating a long and elaborate prompt, crafting their prompt based on the particular TTI type, and describing their use of multiple TTI techniques in combination. 
We compare {\em exploration} based on how users related the \(\displaystyle \estimate\) back to their worldview and experience, and if and how they used the classifier outputs to inform their estimate of the model's worldview, as well as if they observed worldview aspects in the outputs which were not quantified by the classifier. 
We analyze a user's {\em motivation} based on why and how they chose to edit the TTI worldview. We recorded if their motivation was related to a fairness concern, a desire for more diversity or the desire for specific control over image output. 
We compared the {\em analysis} across users recording if the edited outputs reflected a worldview that was closer to their own, and if there were any aspects of a worldview that did not match expectations and why.

\subsubsection{\worldviewer helps users expose how model worldviews align or conflict with their own.} \label{sec: SupportingWorldviews}
Given the diversity of users and their worldviews, the TTI model's worldview did not align with all users.
A common misalignment was between TTI models and users from non-western geographical contexts [P3, P5, P7, P9, P10, P11, P12, P13, P14, P15, P17]\,---\,
\worldviewer revealed that the generated images did not align with their geographical context.
For instance, P14, a professional TTI user, tested a prompt they would regularly use to help them generate architectural renderings: ``\texttt{architectural photograph of alkarnak temple in Luxor egypt, ultrarealistic}''. 
Seeing the model output images of large columns from the temple, P14 considered whether the model reflected a worldview that did not reflect Egyptian architecture or culture.

Revealing misalignment was important for users to understand what a model might be good at (e.g., Modern western architecture but not Egyptian), but it also revealed places to dig deeper and call out inequities.
For instance, when P7 tested the prompt ``\texttt{A female maasai warrior herding their cattle in Nairobi}'', they noticed that, based on their experiences growing up in Kenya, the female warriors were accurately depicted by the Nairobi context was not. 
Instead of a skyline, which is what they expected to see, the background of the images depicted a savanna\,---\,a common type of misalignment they experience in depictions of their home country, which they have come to circumvent using Afro-fusion stylistic techniques as well as blending multiple TTI images.
Similarly, P14 found that TTI models poorly reflect their Egyptian worldview to the point where they were able to generate high-quality images of famous British architect Zaha Hadid, but struggled to generate even something as ubiquitously recognizable as the Egyptian pyramids.
They observed that ``\textit{unfortunately the way Egyptian heritage is being documented is by European colonizers. Because of the worldviews [...] the representation of the heritage in a lot of points is wrong.}''

\worldviewer also supported users whose worldview aligned with the TTI model [P1, P2, P4, P6, P8, P16, P18].
For example, P16 prompted the model to generate ``\texttt{a family dinner}'' and observed that the generated image reflects a family similar to their own.
Seeing this overlap, P16 realized that the model captures aspects of their worldview, and they anticipated having to make only minimal edits to the model's outputs.
Similarly, P6 tested the prompt ``\texttt{a farmer in their field}'' and observed that the farmer's darker skin tones and hats resemble those they are familiar with growing up in New Mexico and California.
Discovering that the model aligned with their worldview made users more aware of the fact the model may not align with other user's worldviews (or aspects of their own worldview, like values of equity and inclusion).
For instance, P16 was spurred to interrogate other aspects of the model's worldview and test how other populations would be represented. They realized how, in their own illustrations, they only include drawings of people that look like them as they are more familiar with their facial features and are hence easier to draw. They found that being able to represent other demographics through TTI would be helpful in amplifying representation in their design repertoire.  
Several other participants [P2,P6,P8] were similarly curious about increasing representation of other groups\,---\,for example, P8 prompted the model to generate ``\texttt{man drinking coffee in a park}'', and edited the guidance to include Black and White male of 20-29 and 70+ year olds to include more diversity in the images.

\subsubsection{Generating diverse TTI outputs requires time and expertise.} \label{sec: TTIworkflow}
Less experienced participants [P2,P3, P6, P12, P16] generated shorter and more generic prompts that did not describe a specific creative vision, providing insufficient  \context for the model, and generally had less intuition for prompt outputs. For example, P6 would create short prompts such as ``\texttt{photo of a person in a pasture}'', without articulating the pose, style of the image, time period or context, and was surprised when the photos depicted modern day people, as opposed to historical photos of people in a pasture which they were expecting to see. Less experienced prompters would often only include the subject in their prompts (e.g., ``\texttt{ a family dinner}''), omitting the type of image (e.g., a photo, a painting, etc.) and were surprised when the model would create nonsensical outputs containing text that read \texttt{``family dinner''} and did not include people in the image.

In contrast, experienced users [P4, P5, P7, P11, P14] wrote longer prompts (i.e., several sentences and even paragraphs) with explicit descriptions of the \context (e.g., human figure poses, action verbs, and lighting). 
To author these prompts, P7 and P14 explained that they incorporate multiple techniques and tools\,---\,for instance, aligning their TTI prompt to the output of an image-to-text model fed with a reference image that they liked.
In addition to prompting techniques, more experienced users also developed their own design exploration workflows, which consisted of first generating a large number of images per prompt and then narrowing it down to a single image that reflected their expectations [P1, P2, P4, P5, P6, P7, P9, P10, P11, P14, P15]. 
Some participants described an iterative process where they used TTI for initial concept visualization, used TTI to modify specific features in the image, and then further refined images through external tools, such as Photoshop~\citep{adobephotoshop} [P1, P4, P5, P7, P14, P15, P16, P17]. 
Ultimately, however, expert users described that, to ensure the models are able to reflect the worldview or stylistic quality they intended, they often needed to fine-tune their own models or use specialized models from the TTI community. 
For example, as P14 described \textit{``[to] create good artwork that represents a certain style you need a model that is 100\% created for that style [...] that has these heritages''}.

\subsubsection{The ideal TTI outputs and editing technique depends on user tasks.}
\label{sec: algo-expectation}
When prompting participants to choose a worldview editing technique to apply, all participants paused and commented on the difficulty of the question and chose different editing techniques over the several prompts they tested. Participants tested several worldviews on different prompts and concurred that choosing a worldview is context dependent. 
4/18 of participants were undecided between parity and US demographics as they were uncertain about whether they should represent an ideal world or the real world. 
P2 was undecided between relative technique and US demographic, as their goal was to match the real world distribution as well as tweak the model to marginally improve representation. 
5/18 users selected US demographics\,---\,as P1 explained when testing the prompt ``\texttt{A photo of a fashion blogger}'', they noticed how the original image represented a narrow spectrum of people, and therefore chose to apply the US demographic edit to mimic the ``\textit{diversity of representations that one sees when walking on the street}''. 
5/18 participants chose absolute editing due to the ease of reaching their design goals\,---\,either having having a specific vision in mind, or aiming to represent a specific minority or themselves. 
For example, P15 tested the prompt ``\texttt{The CEO of a cosmetics company presenting the latest innovation in skincare}'' and noticed that the output images represented all female figures with light skin tones. They applied the absolute edit as a way to gain intuition in \estimating the outputs of the edited model. 
This variety of approaches reinforces the idea that ubiquitous worldviews do not satisfy users; and that context, such as uses of the generated images, determines what worldview users think should be applied to the model. 

\subsubsection{Users wished for editing categories beyond demographics.} \label{sec: fair-categories}
After users tested the editing pipeline, we asked participants whether there were any worldview editing options that they wished were included as well as other considerations that were missed. 
Across the board, we observed that each participant had different ideas about which categories should be included. Some users were satisfied with the gender, race and age categories included in \worldviewer [P10, P11], while others suggested expanding the gender options and racial categories, as well as adding the ability to modify background and context, people's emotions, body type, wealth and clothing. Some users expressed an interest in being able to specify demographic worldviews of different geographic locations and communities within the worldview options [P1,P2, P3, P5, P6, P8, P9, P10]. P1 observed how when they tested the prompt ``\texttt{``A photo of a fashion blogger}'', the outputs of the model were primarily thin, female figures, posing in front of a city background. They suggested that including background editing and body type editing would allow for currently under-represented worldviews, such as their own, to be generated via the model. 
While testing the prompt ``\texttt{``A photo of a suspicious person}'', P2 also commented on the importance of background in their tasks. They are developing a dataset to detect suspicious behavior outside of people's homes, and being able to edit the background of the image to capture different urban and suburban contexts would be important in their tasks.
In contrast, P10 warned about the difficulty in deciding which categories to be included as there could be infinitely many, and always a new one that is left out: ``\textit{If we want to include all kinds of social demographic characteristics there can be infinitely many [...]. How can we ensure that all are included?}''
We believe it is a compelling opportunity for future work to consider how to expand the definition of worldviews\,---\,for instance, as we discuss in \autoref{sec:discussion}, perhaps a method of composing worldviews together might balance the diversity of categories people wished to be able to modify with the tension of needing any worldview to be comprehensively descriptive.

\subsubsection{The potential and danger of modifying image demographics} \label{sec:52}
While many participants believed that editing worldviews can empower minorities through better representation [P1,P2, P3,P4, P6, P7, P8, P9, P10, P11], other participants questioned whether users should have tools to modify the worldviews as they may lead to adversarial and harmful worldviews [P5, P12]. 

Participants in favor of allowing users to modify the worldviews in the image cited the potential for increasing the visibility and representation of minority groups. 
For example, when editing images of ``A photo of a rapper who's won a grammy'', P12 explained how applying parity in this context may be inspirational for rappers who are not currently represented amongst winners, and ``\textit{just because it hasn't happened doesn't mean you shouldn't represent it.}''
Similarly, participants posited that offering users control over model worldviews could increase awareness of model biases: allowing users to modify and analyze the worldviews in the model would empower users to make a conscious decision about which worldview to apply, rather than being able to altogether ignore the underlying worldview. As P8 explains, choosing not to edit a worldview should be a conscious decision:  ``\textit{It’s a choice that people should make, not the machines. [...] 
    A non-choice is a choice. [...]
    A good way forward would be to have these sorts of tools more upfront to be aware of what they want to represent}''\,---\,a sentiment that was echoed by P16.

In contrast, participants who opposed offering options to modify worldviews were concerned about potential adversarial uses of the tool and users with ill-intentions. 
For example, when prompting the model to generate a ``\texttt{A photo of a rapper who's won a grammy}'', P12 worried about the ways these tools could be used to generate specific demographics in ways that represent minorities in a bad light or in a discriminatory fashion:

\begin{quote}
``\textit{[...] maybe there are people out there that hold very bad worldviews of minorities and would just generate specific kinds of images to portray them in a certain kind of way that is promoting discriminatory views. They may try to generate toxic images about minorities in an adversarial way.}''~--~P12
\end{quote}

Similarly, P5 worried that brands could have bad-faith motives to modify images of models to appear more diverse, without actually diversifying their workforce:

\begin{quote}
    ``\textit{I have reservations to choosing, diversity. This goes back to the idea of why do we want diversity, if it is to make changes in the world. You could make the argument that images end up shaping the world we live in, but you could also make the argument that that is not entirely true. It is designed for marketing, for image making. It is not necessarily designed to generate an honest representation. [...] Honesty is very important. [...] it depends on which person is using that data.}''~--~P5
\end{quote}

\subsection{Limitations of \worldviewer}
\label{sec:user-study-limitations}

\subsubsection{Improving the demographic classifier accuracy and categories}
Throughout the interviews, users noticed discrepancies between their visual perception of the demographics represented in the generated images and the classifier's predictions. While discrepancies in age and gender classifications were not as apparent, race predictions were obvious and misleading. Errors in classification could be related to the fact that the classifier is trained on real photos and not generated images. 
While the majority of users noticed issues with the accuracy of the classifier, they relied on their visual perception of the demographics in the generated images to deem if the results were fair, therefore they did not hinder users workflows. One participant (P13) who self-identified as face blind, however, relied more heavily than other participants on classifier information to determine the demographics of the outputs. They expressed how having a classifier to detect demographics was useful in their tasks.   

\subsubsection{Semantic editing does not work in every \context, rather it highlights that biases cannot be easily removed}
While participants noticed an improvement in representation across edited images, several users pointed to the decrease in image quality of the people in edited images. The decrease in quality of the figures, as well as the inability to generate sensical images in some \context (\cref{fig: UserStudySurgeon}), further reinforced the notion that editing is not always possible, and pointed to limitations in moving through latent space and achieve their distributional goals via \( \texttt{semantic\_guidance} \).   %
From the interviews it is clear that there can be a disconnect between user expectations and algorithmic implementation. This can occur when people building the algorithm are disjointed from the end user expectation. 
Several users comment on how parity or the US demographic editing techniques don't generate the exact representation they were expecting to see [P1, P5, P6, P9, P10, P12]. P6 notices the misalignment and inquires on the \( \texttt{semantic\_guidance} \) \concept assignment. They suggest that hard coding edits may allow to generate distributions which more closely represent the US demographics and parity than through random coin flips, given the limited number of output images [P6]. 
Limitations in image quality point to the downstream effects of \capta on learned \arrows, and to the constraints in modifying worldviews once a model is trained: re-training or fine-tuning may be necessary to achieve under-represented worldviews.

\section{Evaluation: Case Studies}\label{sec:casestudy-section}
We illustrate how \worldviewer allows to improve the representation in TTI models through two case studies of common TTI workflows drawn from users in our user study (\cref{sec: userstudy}).

\subsection{Representing Diverse Skin Tones} 
\begin{figure*}
   \includegraphics[width=\linewidth]{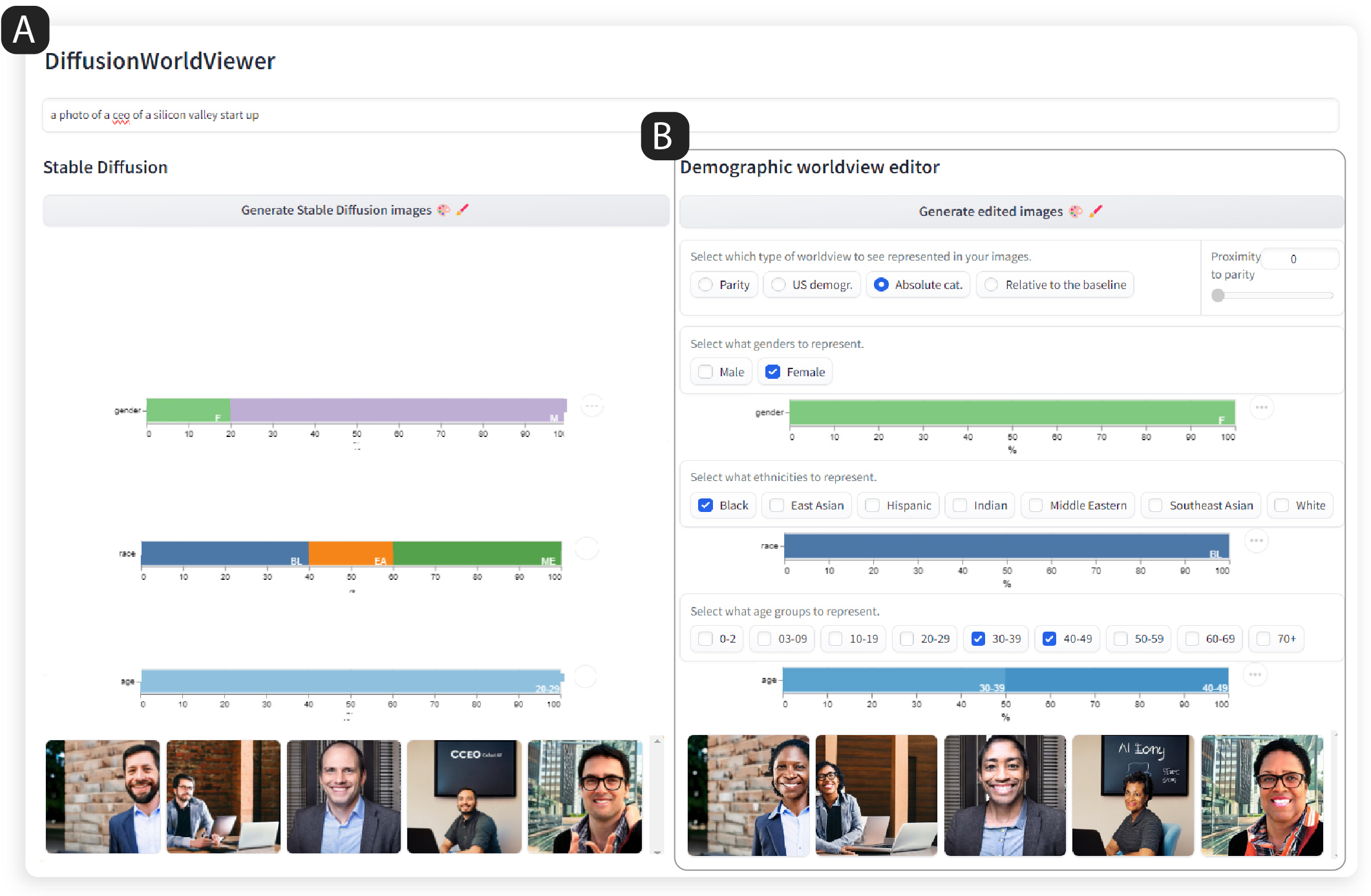}
   \caption{\worldviewer applied to the case study Minority representation in professional context compare s Generative Text-to-Image model representations for the prompt ``a photo of a ceo of a silicon valley start up''. (A) The baseline images generated represent male, middle aged with light color skin. (B) The user applies the absolute category to represent CEOs that look like them, selecting the female, black and 30-39 and 40-49 year old individuals. The edited images all represent black individuals. Images 2,4, and 5 represent female individuals, while images 1 and 3 are ambiguous.}
  \label{fig: CaseStudy2}
  \Description{\worldviewer user interface with 2 column design. The page consists of a text box for users to type their prompts at the top where a user has typed in "a photo of a CEO silicon valley start up", followed by a main area subdivided into a left and right hand side. On the left hand side(labelled A), is the title "Stable Diffusion" followed vertically by the button, which can be pressed to "generate baseline images", followed vertically by 3 single stacked bar charts- the top one showcases gender, which depicts 20\% females and 80\% males, the middle one showcases race, which depicts 40\% Black, 20\% east asian and 40\% middle eastern individuals, and the bottom one showcases age groups, which depicts 100\%20-29 year olds. There is an image gallery, which depicts a row of 5 images containing representations of light skin tone and caucasian looking individuals. 
   On the right hand side (labelled B) there is the title "Demographic worldview editor", followed by a button, which can be pressed to "generate edited images", followed by a 4 worldview buttons "Parity", "US demogr.", "Absolute cat.", "Relative to the baseline". In the image the button "Absolute cat." is selected. To the right of the worldview buttons is a slider titled "Proximity to Parity", which is greyed out. Below there are 2 types of components that are alternated: a group of checkboxes where users can "Select what genders to represent" and can choose between "Male" and "Female" - "Female" is selected ; a single stacked bar chart representing selected gender distributions which reflects 100\% females; a group of checkboxes where users can "Select what ethnicities to represent" and can choose between "Black", "East Asian", "Hispanic", "Indian", "Middle Eastern", "Southeast Asian" and "White" - "Black" is selected ; a single stacked bar chart representing selected ethnicity distributions which reflects 100\% Blacks; a group of checkboxes where users can "Select what age groups to represent" and can choose between "0-2", "3-9", "10-19", "20-29", "30-39", "40-49", "50-59", "60-69", "70+" - 30-39 and 40-49 are selected; a single stacked bar chart representing selected age group distributions which reflects 50\% 30-39 and 50\% 40-49. Below that is an image gallery, which depicts a row of 5 images containing representations of middle-aged, darker skin tone individuals of different genders.}
\end{figure*}

Inspired by conversation with P7, this case study demonstrates how \worldviewer can be used to improve the representation of minorities in contexts where they typically lack representation. 
Here the user who grew up in Kenya, explains how in the past when they tried making birthday posters and art of their friends using TTI, they had to stop due to the difficulty of representing darker skin tones and facial features accurately.

The user, a business professional and aspiring CEO, utilizes the baseline Stable Diffusion model to prompt ``\texttt{a photo of a CEO of a silicon valley start up}'' and generates 5 images. 
The user explains how, based on their experience using TTI and their experience in the entrepreneurship space, they expect to see images of white males giving a presentation since they are historically most represented. 
They are interested in seeing if they can use the tool to represent their unique perspective, and generate CEOs that also have darker skin tone. 

The user selects the absolute editing technique to see if they can generate people that look similar to themselves in the images. 
They select to represent female gender and Black ethnicity, as well as 30-39 year olds and 40-49 year olds, and generate 5 new images (Figure \ref{fig: CaseStudy2}B).
The user scans the images and observes that the individuals look like individuals of color. They zoom into the images, and observe that figures represented 2, 3 and 5 are seemingly female, while 1 and 3 are unclear. 

The researcher was excited about the results and representation the \worldviewer helped reveal.
Generating this type of representation with prior tools required significant tedious effort or was not possible to achieve. 
Previously, the researcher would have needed to iteratively construct and test prompts to generate a specific demographic, which would have required several tries and compute time to achieve: 
\begin{quote}
``\textit{this [tool] is going to make my work a lot easier.[...] A lot of text to image prompts are very generic and there is a lot of hit or miss [...]. This puts an immense amount of control.}''~--~P7
\end{quote}

In contrast, by selecting demographics to represent, the \worldviewer surfaces this information more directly, without leaving a user to guess which demographic will be generated by default by the model. 
They explain that this technique \textit{``is a game changer''} for them, since previously they were not able to represent themselves and their friends well using TTI, and will now be able to use these tool to make art for their friends and family. 
As a result, the system transforms the process of testing and comparing prompt outputs, towards more of an introspective choice of whom to represent. 
This shift frees the researcher to test and think about which worldview to apply, rather than focusing effort on iteratively prompt engineer to achieve a specific demographic representation. 

\subsection{Representing Community}
Inspired by conversation with designer and researcher P9, this case study demonstrates the \worldviewer used in an illustration setting. 
The user's goal is to generate images that are representative of the community they are preparing a presentation for: children in STEM. 
Their goal is to represent a wide range of ethnicities of kids in STEM. They are concerned that images will not represent the wide variety of backgrounds, so they are curious to see if they will be able to represent their target community. 
\begin{figure*}
   \includegraphics[width=\linewidth]{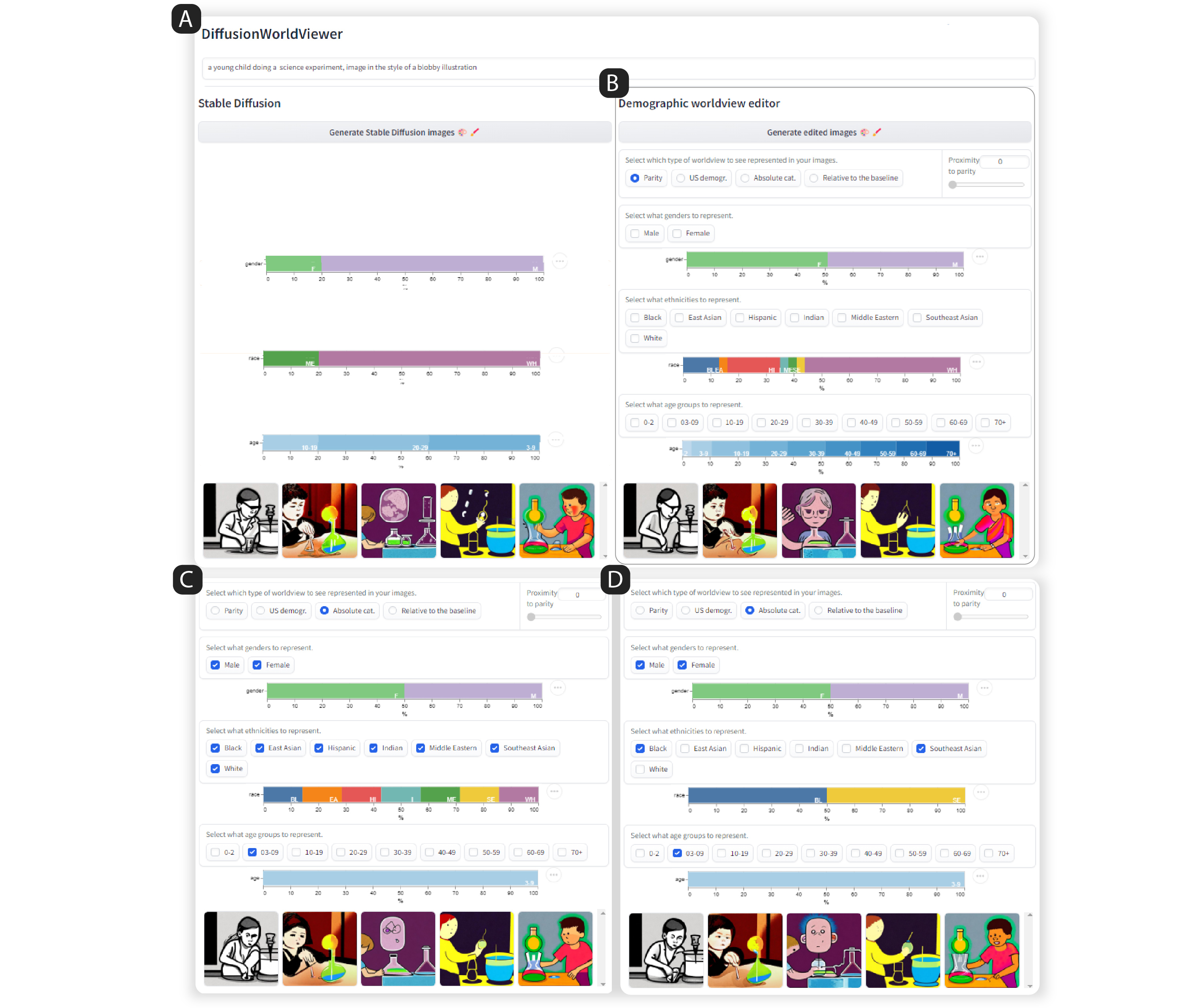}
   \caption{\worldviewer applied to the case study Representing Community explores TTI model representations of ``\texttt{a young child doing a science experiment, image in the style of a blobby illustration}'' (A). The user explores different worldview editing techniques including, parity (B) and absolute editing (C--D) to generate illustrations that represent their audience community. }
   \Description{A \worldviewer user interface with 2 column design, followed by 2 screenshots only of the editing pipeline. The interface consists of a text box for users to type their prompts at the top where a user has typed in "a young child doing a science experiment, image in the style of a blobby illustration", followed by a main area subdivided into a left and right hand side. On the left hand side(labelled A), is the title "Stable Diffusion" followed vertically by the button, which can be pressed to "generate baseline images", followed vertically by 3 single stacked bar charts- the top one showcases gender, which depicts 20\% females and 80\% males, the middle one showcases race, which depicts 20\% Middle Eastern, 80\% White, and the bottom one showcases age groups, which depicts 20\%10-19, 40\%20-29 and 40\%3-9 year olds. There is an image gallery, which depicts a row of 5 images containing illustrations of light skin tone boys doing science experiments. 
   On the right hand side (labelled B) there is the title "Demographic worldview editor", followed by a button, which can be pressed to "generate edited images", followed by a 4 worldview buttons "Parity", "US demogr.", "Absolute cat.", "Relative to the baseline".  In the image the button "Parity is selected".  To the right of the worldview buttons is a slider titled "Proximity to Parity", which is greyed out. Below is a button, which can be pressed to "generate edited images". Below there are 2 types of components that are alternated: a group of checkboxes where users can "Select what genders to represent" and can choose between "Male" and "Female"; a single stacked bar chart representing selected gender distributions which is half male and half female ; a group of checkboxes where users can "Select what ethnicities to represent" and can choose between "Black", "East Asian", "Hispanic", "Indian", "Middle Eastern", "Southeast Asian" and "White" ; a single stacked bar chart representing selected ethnicity distributions which are equally proportional amongst all ethnicities ; a group of checkboxes where users can "Select what age groups to represent" and can choose between "0-2", "3-9", "10-19", "20-29", "30-39", "40-49", "50-59", "60-69", "70+" ; a single stacked bar chart representing selected age group distributions which are equally proportional amongst all ages . Below that is  an image gallery(labelled L), which depicts a row of 5 images containing more diverse representations of gender and age, while skin tone and composition match the original images.
    The first editing pipeline screenshot (labelled C) depicts the selection of Absolute editing category, where all boxes are selected across gender and ethnicities, and the 3-9 year old is the only selected category for age. Images in the gallery maintain the same composition and skin tones, and images are reverted to depict young children. 
   The second editing pipeline screenshot (labelled D) depicts the selection of Absolute editing category, where all boxes are selected across gender, "Black" and "South East Asian" are selected across ethnicities, and the 3-9 year old is the only selected category for age. Images in the gallery maintain the same composition, while the 3rd and last image skin tones change and gender becomes ambiguous. }
  \label{fig: CaseStudy3}
\end{figure*}

At first, they test the prompt ``\texttt{a young child doing a science experiment, image in the style of a blobby illustration}''.
They are curious to see which genders will be represented in the illustrations. They notice that out of the 5 generated images, all 5 images resemble five young boys and 3 out of the 5 images appear to be light skinned. 
The ethnicity of the remaining 2 images are unclear as the stylistic drawing and colors, are ambiguous and hard to interpret (\cref{fig: CaseStudy3}A).

By glancing over at the editor on the right, the user notices the parity option. 
This option shows equal proportion of male and female, as well as ethnicities, so they test this option and generate 5 new images. 
Looking at the 5 new images, the user can see that the new images resemble the previous ones. 
They notice that there are 2 images that seem to represent girls, 2 that seem to represent boys, and 1 that is gender ambiguous (\cref{fig: CaseStudy3}B). 
The images are more representative, but one of the images has changed to an older aged women, which was not their intent.

To have more fine-grained controls, they decide to use the absolute editing feature, and check boxes to represent all genders and races, and 3-9 year olds. 
They generate images again (\cref{fig: CaseStudy3}C) and notice that the child in the second image is more feminine, the third image has reverted to the original image, the fourth is still ambiguous, and the final image looks more masculine.

Now they are curious to see if they can explicitly represent change all of the images to represent Black and South East Asian children that they know will be present at their presentation.
They generate 5 more images (\cref{fig: CaseStudy3}D) and notice that figure 2 and 5 represent South East Asian children, but notice how the other images are not as successful. 
Image 1 seems to represent an Asian person, but they seem older than 9 years old. 
Image 3 seems to have edits the smoke from the beaker into the head of a person, and image 4 is remains ambiguous.

The \worldviewer allows designers and researchers to compare alternate worldviews, which is valuable in this case as user's expectation and model outcome don't always match. 
Testing and comparing the outcomes and demographic plots of parity, and absolute category, the user finds that parity is enforced across race, gender and  age, which goes against their intent of generating children in their image (\cref{fig: CaseStudy3}B). 
By exploring other worldviews and fixing gender and race to parity, while selectively choosing only specific ages through absolute editing, the researcher is able to more closely match their expectations (\cref{fig: CaseStudy3}C).

\section{Discussion and Future Work}
\label{sec:discussion}
In this paper, we present \worldviewer, an interactive tool to assess and edit the worldviews of generative text-to-image (TTI) models.
\worldviewer surfaces the demographic distributions of the age, race, and gender of TTI generated images, revealing cases where a TTI model's worldview may be misaligned with the user's expectations of diversity.
In cases of misalignment, \worldviewer allows users to directly edit generated images via semantic guidance by shifting the outputs towards the user's ideal demographic distributions.
To analyze the design of the system and reveal how practitioners think about the worldviews represented in images, we conducted user studies with 18 TTI model users with diverse geographical and cultural backgrounds, demographics, experience levels, and TTI use cases.
Our user studies reveal how \worldviewer encourages conversations about representation in technology by directly exposing otherwise hidden TTI model worldviews and enables users to reflect their worldviews in generated images through output editing tools.
Further, through case studies of \worldviewer applied to two common TTI tasks, we show that \worldviewer shifts TTI use from tedious prompt-based searching to efficient and effective edits.

We designed \worldviewer to enable users to generate images via prompting and edit the generated outputs using semantic guidance.
However, an inherent trade-off exists between these two methods.
While semantic guidance offers personalized edits that reflect a worldview beyond the TTI model's prompt-generated outputs, if a prompt exists that could generate the user's ideal output, editing may produce lower quality outputs.
This has been validated experimentally by \citet{gonen2019lipstick} in word embeddings and anecdotally by our users (\cref{sec:user-study-limitations}).
The nuances involved in determining the most effective way to shift the model's worldview underscore the importance of communicating TTI editing trade-offs to users, especially those lacking a machine learning background. 
Although we intentionally designed \worldviewer as a code-free user interface and our study reveals that users without ML expertise or prior TTI experience were able to leverage \worldviewer to generate content that aligned with their worldview (\cref{sec: userstudy}), it does not explicitly teach TTI model literacy skills.
As generative models become increasingly prevalent and are employed by lay-individuals in their day-to-day tasks, future studies should broaden the focus to include a larger set of novice users (e.g., teachers and the elderly) to understanding how worldview editing tools and TTI model interfaces can support learning and intuition building.

A key advantage of \worldviewer is its ability to empower users to edit generated images, allowing for the expression of their unique worldviews that may not be achievable by the original TTI model, such blobby-style images of racially diverse children (\cref{sec:casestudy-section}).
Nevertheless, the manipulation of image demographics introduces ethical considerations.
While users may leverage \worldviewer to increase image diversity, reflect the demographics of their task, and expand the worldview of the original TTI model, an adversarial user could exploit it to further restrict the TTI model's worldview to produce homogeneous images even when working on a task that calls for broad representation.
Worse, there is potential risk of editing tools increasing fairness gerrymandering, where users can easily generate `fair' images without needing to contribute to the diversity of their community or the TTI technology.
In our user study, we found that exposing the demographics of the TTI model broadens users perspectives of diversity by identifying where the TTI model's worldview is limited (e.g.,~\cref{sec:52} and ~\cref{sec: SupportingWorldviews}) and where it may be more broad than their own (\cref{sec: SupportingWorldviews}).
However, as TTI models and their editing tools expand, so should conversations about ethical technology usage and research into methods and systems that educate against adversarial usage, such as automated suggestions in the UI that could prompt a user to expand the worldview of the generated images or methodological guardrails against preventing explicitly harmful edits based on the prompt.

Similarly, while \worldviewer and TTI output editing algorithms offer users the ability to efficiently tailor outputs to reflect their worldviews, it is important to recognize that these tools are not substitutes for developing TTI models that represent broad perspectives.
Instead, worldview tools and TTI model development should be viewed as co-adaptive technologies.
By revealing the TTI model's worldview, tools like \worldviewer can serve as valuable indicators of the model's limitations and guide targeted model development efforts.
For instance, using \worldviewer our user study participants identified that Stable Diffusion predominantly reflects a white and Westernized worldview as well as more subtle limitations, like its preference towards slim body types and modern architecture (\cref{sec: userstudy}).
Seeing these limitations spurred users to generate hypotheses for how to develop models with expanded worldviews, including fine-tuning TTI models on culture-specific datasets and blending multiple outputs into a single image.
Given even users without ML backgrounds were able to ideate model development strategies suggests that model developers may also benefit from worldview editing tools, like \worldviewer, to debug the worldviews of their models before deployment and identify ways to expand their worldviews by gathering targeted \capta and enforcing particular \arrows through training constraints.
Encouragingly, research has begun crafting customized TTI models for specific demographics~\citep{japanese_stable_diffusion, indigenous2023poc}, and we anticipate exposing additional worldview limitations of TTI models will catalyze additional models and methods.

Nevertheless, worldview editing tools are still needed to reflect the individuality of user worldviews.
Even as TTI models evolve to encompass more diverse perspectives, it is improbable that a single model or set of models will capture every facet of global culture, an individual's unique perspective, or the needs of every user task.
While \worldviewer explored demographic editing as a way to help users incorporate their values on representation and the needs of their task in TTI generated images, we envision its contribution as laying the groundwork for future research into identifying the requisite set of axis that are minimally sufficient to reflect user worldviews and ways to incorporate user customizable edits into the semantic guidance methodology and UI.
Our current approach relied on discrete demographic descriptions and notions of parity to satisfy semantic guidance's requirement of a discrete concept; however, this does not represent the continuous and multi-dimensional nature of gender and race, can not categorically represent many minority groups, and does not represent the range of fairness definitions a user may hold.
At the same time, worldviews are nebulous\,---\,they can not be synthesized and categorized into a set of editable parameters.
As a result, future work may study how to incorporate these continuous variables into editing algorithms, converting abstract ideas of a worldview into concrete concepts that could be used in semantic guidance, and surfacing a multitude of fairness definitions beyond parity.

Further, while \worldviewer studied how individuals can understand a TTI model's worldview and utilize prompting and editing to reflect their worldviews, real-world users are often tasked with balancing the worldviews of multiple individuals from diverse backgrounds, such as mixed-skill teams or diverse communities.
In these contexts, interactive user interfaces emerge as critical tools for expressing and amalgamating edits to generate outputs that reflect the combined interests of many worldviews. 
By leveraging our \worldview formalism (\cref{sec:worldview-formalism}), we can structure the design of these tools by conceptualizing them as functions operating on a \worldview.
This formalism allows us to consider designing function that compose worldviews from multiple users ($\worldview + \worldview$), exclude a particular worldview from the outputs ($\worldview - \worldview$), or enhance one worldview by another ($\worldview * \worldview$).
Such a formalized approach offers a systematic framework for expanding \worldviewer and designing future worldview editing tools that support all users in comprehending TTI model worldviews and effectively reflecting their diverse perspectives in generated outputs.

\begin{acks}
This material is based upon work supported by the National Science Foundation Graduate Research Fellowship under Grant No. 2141064. Any opinion, findings, and conclusions or recommendations expressed in this material are those of the authors(s) and do not necessarily reflect the views of the National Science Foundation.
\end{acks}

\bibliographystyle{ACM-Reference-Format}
\bibliography{bib}

\newpage
\onecolumn
\appendix{
\section{Appendix}

\subsection{Additional Case Studies}

\subsection{Inclusive representations for marketing material}

The first case study demonstrates the \worldviewer used to generate marketing material for a retirement home. The designer is tasked with creating depictions of the home without infringing on current resident's privacy. Therefore, they resort to generative image models. They use the baseline Stable Diffusion model to generate 5 images for the prompt ``a marketing photo of a happy retirement home'' (Figure \ref{fig: CaseStudy1}A). They notice that images prevalently depict females and want to improve the gender distribution across images. They choose to edit their images with US demographics, as they expect this option to better capture the diversity of people living at the home (Figure \ref{fig: CaseStudy1}B). They generate 5 more images, and notice that the background in the images stayed the same. The posing in image 1, 2, and 5 has changed, and ethnicities in image 2, 4, and 5 have seemingly changed to asian and black.

\begin{figure*}
   \includegraphics[width=\linewidth]{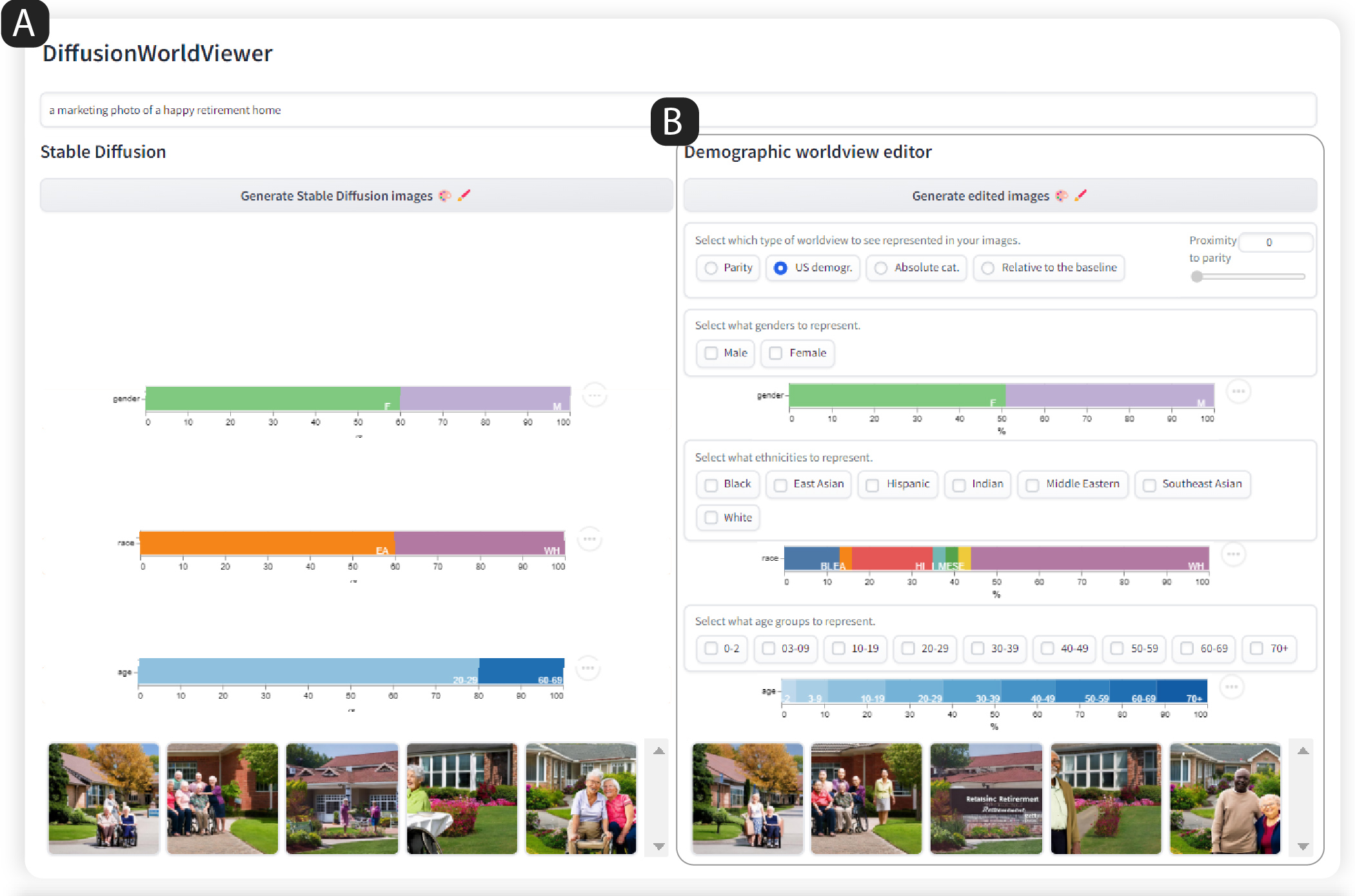}
   \caption{\worldviewer applied to the case study \em{Inclusive representations for marketing material} compares Generative Text-to-Image outputs of the prompt ``a marketing photo of a happy retirement home''. (A) The baseline images reveal the underlying worldview of the model - people represented by the model are prevalently white, old with grey colored hair. (B) Selecting the U.S. demographics editing technique, modifies the images to include seemingly asian and black ethnicities in figures 2, 4, and 5. }
   \Description{\worldviewer user interface with 2 column design. The page consists of a text box for users to type their prompts at the top where a user has typed in "a marketing photo of a happy retirement home", followed by a main area subdivided into a left and right hand side. On the left hand side(labelled A), is the title "Stable Diffusion" followed vertically by the button, which can be pressed to "generate baseline images", followed vertically by 3 single stacked bar charts (labelled B1-3)- the top one showcases gender, which depicts 60\% females and 40\% males, the middle one showcases race,which depicts 60\% east asian and 40\% white individuals, and the bottom one showcases age groups, which depicts 80\%20-29 and 20\% 60-69 year olds. There is an image gallery, which depicts a row of 5 images containing representations of light skin tone, white haired individuals sitting in front of residences. 
   On the right hand side (labelled B) there is the title "Demographic worldview editor", followed by a button, which can be pressed to "generate edited images", followed by a 4 worldview buttons "Parity", "US demogr.", "Absolute cat.", "Relative to the baseline". In the image the button "US demogr." is selected. To the right of the worldview buttons is a slider titled "Proximity to Parity", which is greyed out. Below there are 2 types of components that are alternated: a group of checkboxes where users can "Select what genders to represent" and can choose between "Male" and "Female" ; a single stacked bar chart representing selected gender distributions which reflect US population ; a group of checkboxes where users can "Select what ethnicities to represent" and can choose between "Black", "East Asian", "Hispanic", "Indian", "Middle Eastern", "Southeast Asian" and "White" ; a single stacked bar chart representing selected ethnicity distributions which reflect US population; a group of checkboxes where users can "Select what age groups to represent" and can choose between ``0--2'', ``3--9'', ``10--19'', ``20--29'', ``30--39'', ``40--49'', ``50--59'', ``60--69'', ``70+'' (labelled G3); a single stacked bar chart representing selected age group distributions which reflect US population. Below that is an image gallery, which depicts a row of 5 images containing representations of diverse skin tone, gender and old individuals outside of a residence.}
  \label{fig: CaseStudy1}
\end{figure*}

\subsubsection{More representative scale figures using \worldviewer}

The first case study is inspired by power users interviews in the user studies, who describe using reverse image to prompt searching tools to generate natural language descriptions for their prompts, as well as long prompts to minutely describe what they want to see output by the model. Here we showcase a typical workflow of a designer, who's goal is to generate a futuristic rendering and concept art for a new building they are designing. The power user first begins by searching online for reference images, that resemble their design goals. Once they have narrowed down their design references to a single image, they want to find the keywords and prompt that could have generated that image. They use the online img2prompt tool \cite{img2prompt}, to determine a possible prompt that could have generated the image (Figure \ref{fig: CaseStudy4}, A), and paste the prompt ``a large building with a bunch of windows on top of it, a surrealist sculpture by Gaudi, featured on dribble, art nouveau, made of wrought iron, biomorphic, made of vines'' into their Diffusion model prompt bar to generate new images (Figure \ref{fig: CaseStudy4}, B). The generated images resembles the style of the reference image, but now the artist wants to include a human figure walking down the sidewalk. They add the prompt ``a sidewalk below the building, woman walking on the sidewalk'' at the end of the original description. They generate new images using the edited prompt (Figure \ref{fig: CaseStudy4}, C), and observe how 2 of the 5 images have included a person walking: the first image shows a woman wearing a dress and walking on the sidewalk, however the background is the least appealing out of the five. The second image shows a person walking on the sidewalk and more closely resembled what the designer envisions. At this point the designer, notices that the figures both have light skin tones, which do not match the demographics of the neighborhood where the new building will be placed, and wants to better reflect the community living there. Hence, the designer once again edits the prompt to explicitly specify the age and race of the person, by changing the prompt to ``young woman of color walking on the sidewalk''. The user skims over the 5 images and notices that the images have changed substantially and that the composition of some images have changed compared to the original images (Figure \ref{fig: CaseStudy4}, D). The person in the first image, has been distorted into a tree, whilst the third image, which previously did not include photos of a person now has a wider angle to include the sidewalk. Comparing the results from the 2 and 3 set of images, the designer notices the influence of the keywords ``young'' and ``of color'' added to this prompt, and how they have in some instances changed the composition in the images.
\begin{figure*}
   \includegraphics[width=\linewidth]{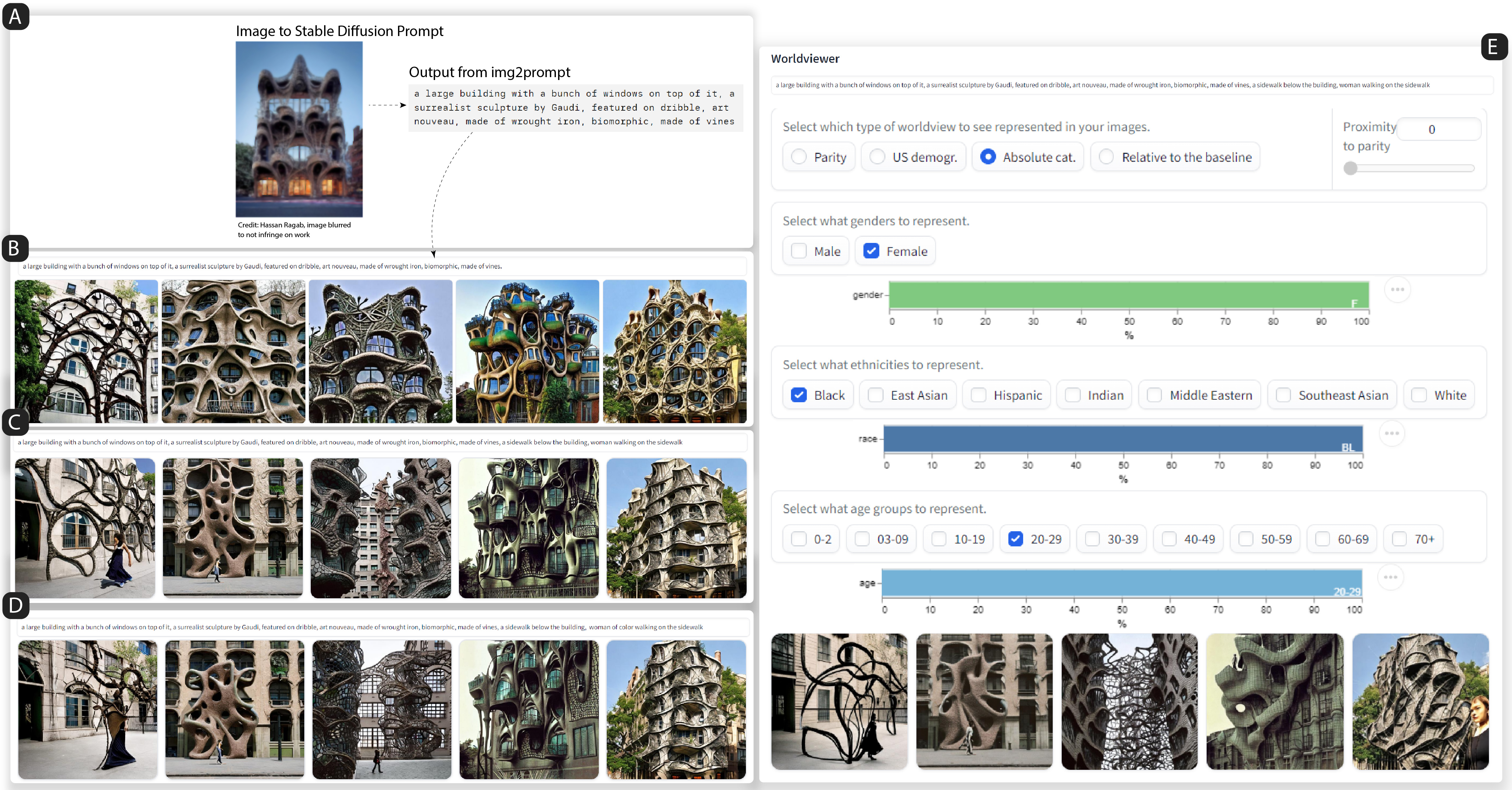}
   \caption{\worldviewer applied to generate more representative human scale figures in architectural design renders. A representative case study depicts an expert user generating a prompt based on a reference image (A) through text2prompt, and aiming to include scale figures that are representative of the neighborhood demographics where the building will be built, using prompt engineering (B-C), and using \worldviewer (D).
   The user tests the prompt ``a large building with a bunch of windows on top of it, a surrealist sculpture by Gaudi, featured on dribble, art nouveau, made of wrought iron, biomorphic, made of vines, a sidewalk below the building,  woman walking on the sidewalk''. }
   \Description{ 2 column diagram representing a user pipeline. 
   The first image (A) represents how a reference image is used to generate the text descriptions for the image using image2prompt tool "a large building with a bunch of windows on top of it, a surrealist sculpture by Gaudi, featured on dribble, art nouveau made of wrought iron, biomorphic, made of vines". The images below show 
   The following images (labelled B,C,D) show how inputting this prompt into the baseline worldviewer generates 5 images of biomorphic looking buildings that resemble the original reference image. Prompt in image B is not modified, while prompt in image C is edited to include "woman walking on the sidewalk" and image in prompt D is edited to include "woman of color walking down the sidewalk". 
   Image E is a screenshot of the editing pipeleine from \worldviewer, where the "Absolute worldview" is selected, "Female", "Black", and "20-29" categories are selected.
   Images C contain a woman walking down the street in 2/5 images. Images D contain a morphed representation of a person in image 1, and people walking in image 2 and 3. 
   Images E contain a morphed representation of a person in image 1, and people walking in image 2.}
  \label{fig: CaseStudy4}
\end{figure*}
The designer notices the \worldviewer dashboard and are curious to see if it will be able to modify the demographics of their images, without distorting the composition of the image. They return to prompt 2, ``a large building with a bunch of windows on top of it, a surrealist sculpture by Gaudi, featured on dribble, art nouveau, made of wrought iron, biomorphic, made of vines, a sidewalk below the building,  woman walking on the sidewalk''. They notice 4 editing techniques, and after some deliberation, decide to apply absolute editing to generate their original vision, of representing a young woman of color walking in front of a futuristic looking building (Figure \ref{fig: CaseStudy4}, E). In the editing dashboard they select the checkbox ``Female'', ``Black'' and ``20-29''.  As images are generated, the user notices how the \worldviewer edited images appear more dull than the prompt edited images. Whilst the figure in the second image seems to have changed in skin tone, the ornamentation of the buildings in the newly generated images is of lower resolution.

\subsection{Additional Figures}
\begin{figure*}
   \includegraphics[width=\linewidth]{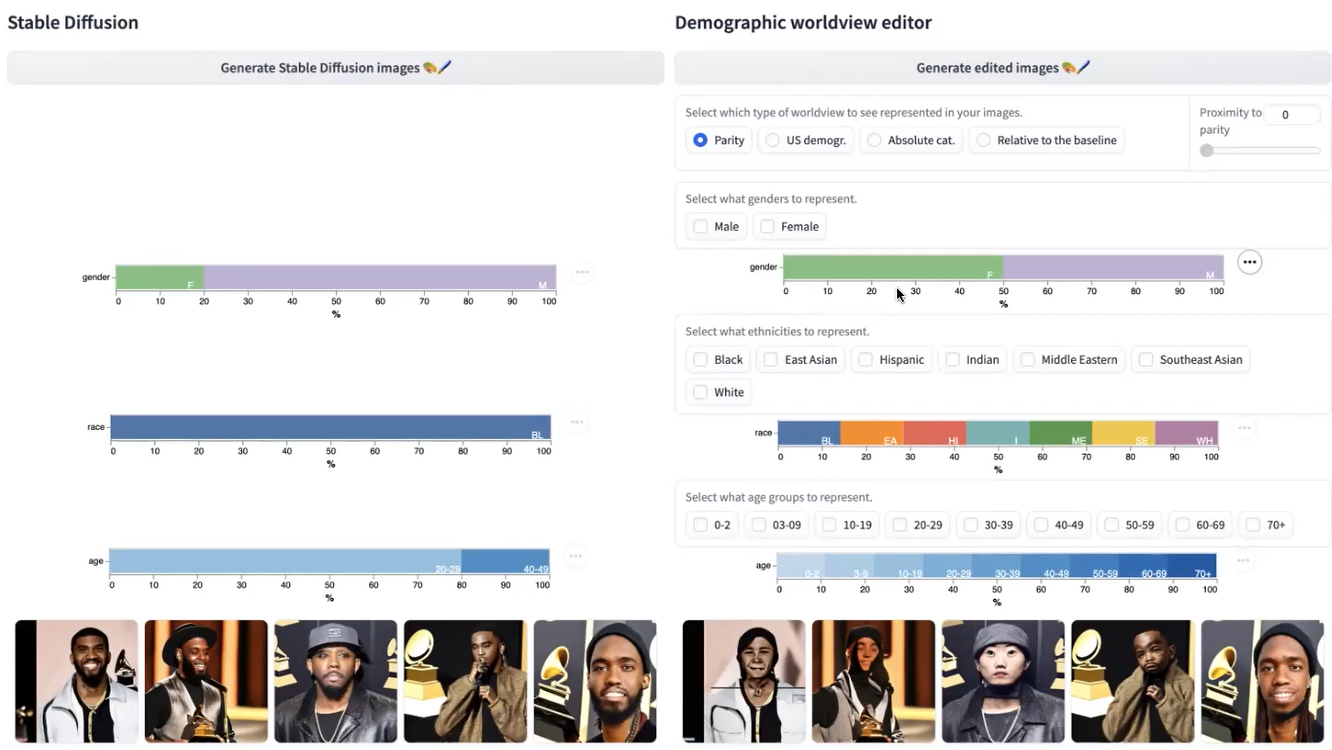}
   \caption{A screenshot from a user study with participant 12, testing the prompt ``A photo of a rapper who's won a grammy'' edited by applying parity. The user explains how applying parity and showcasing diversity could be inspiring for rappers of different backgrounds.}
    \Description{A \worldviewer user interface with 2 column design. The interface is divided into a left and right hand side. On the left hand side(labelled A), is the title "Stable Diffusion" followed vertically by the button, which can be pressed to "generate baseline images", followed vertically by 3 single stacked bar charts- the top one showcases gender, which depicts 20\% females and 80\% males, the middle one showcases race, which depicts 100\% Black, and the bottom one showcases age groups, which depicts 20\%40-49 and 80\%20-29 year olds. There is an image gallery, which depicts a row of 5 images containing illustrations of dark skin tone men holding or standing in front of a Grammy. 
   On the right hand side (labelled B) there is the title "Demographic worldview editor", followed by a button, which can be pressed to "generate edited images", followed by a 4 worldview buttons "Parity", "US demogr.", "Absolute cat.", "Relative to the baseline".  In the image the button "Parity is selected".  To the right of the worldview buttons is a slider titled "Proximity to Parity", which is greyed out. Below is a button, which can be pressed to "generate edited images". Below there are 2 types of components that are alternated: a group of checkboxes where users can "Select what genders to represent" and can choose between "Male" and "Female"; a single stacked bar chart representing selected gender distributions which is half male and half female ; a group of checkboxes where users can "Select what ethnicities to represent" and can choose between "Black", "East Asian", "Hispanic", "Indian", "Middle Eastern", "Southeast Asian" and "White" ; a single stacked bar chart representing selected ethnicity distributions which are equally proportional amongst all ethnicities ; a group of checkboxes where users can "Select what age groups to represent" and can choose between "0-2", "3-9", "10-19", "20-29", "30-39", "40-49", "50-59", "60-69", "70+" ; a single stacked bar chart representing selected age group distributions which are equally proportional amongst all ages . Below that is  an image gallery, which depicts a row of 5 images containing seemingly 2 black men, 1 asian individual ad 2 individuals of ambiguous races. Figure's faces are distorted and do not look human}
  \label{fig: UserStudyRapper}
\end{figure*}

\begin{figure*}
   \includegraphics[width=\linewidth]{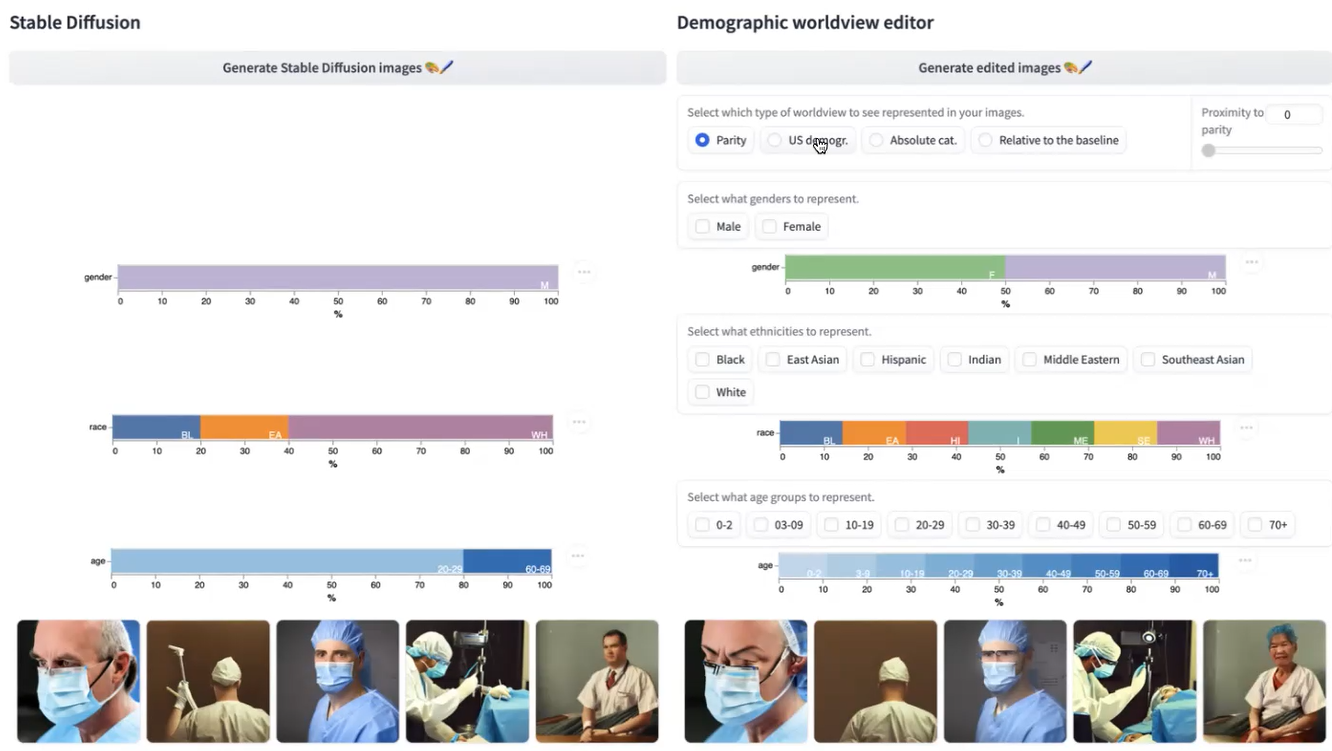}
   \caption{A screenshot from interview with participant 12, testing the prompt ``A photo of a surgeon'' edited by applying parity. The user explains how the similarity between baseline and edited images could point to the tightness and lack of diversity in latent space around the concept surgeon.}
   \Description{A \worldviewer user interface with 2 column design. The interface is divided into a left and right hand side. On the left hand side(labelled A), is the title "Stable Diffusion" followed vertically by the button, which can be pressed to "generate baseline images", followed vertically by 3 single stacked bar charts- the top one showcases gender, which depicts 100\%  males, the middle one showcases race, which depicts 20\% Black, 20\% East Asian and 60\% White, and the bottom one showcases age groups, which depicts 20\%60-69 and 80\%20-29 year olds. There is an image gallery, which depicts a row of 5 images containing illustrations of light skin tone doctors in different poses. 
   On the right hand side (labelled B) there is the title "Demographic worldview editor", followed by a button, which can be pressed to "generate edited images", followed by a 4 worldview buttons "Parity", "US demogr.", "Absolute cat.", "Relative to the baseline".  In the image the button "Parity is selected".  To the right of the worldview buttons is a slider titled "Proximity to Parity", which is greyed out. Below is a button, which can be pressed to "generate edited images". Below there are 2 types of components that are alternated: a group of checkboxes where users can "Select what genders to represent" and can choose between "Male" and "Female"; a single stacked bar chart representing selected gender distributions which is half male and half female ; a group of checkboxes where users can "Select what ethnicities to represent" and can choose between "Black", "East Asian", "Hispanic", "Indian", "Middle Eastern", "Southeast Asian" and "White" ; a single stacked bar chart representing selected ethnicity distributions which are equally proportional amongst all ethnicities ; a group of checkboxes where users can "Select what age groups to represent" and can choose between "0-2", "3-9", "10-19", "20-29", "30-39", "40-49", "50-59", "60-69", "70+" ; a single stacked bar chart representing selected age group distributions which are equally proportional amongst all ages . Below that is  an image gallery, which depicts a row of 5 images which almost identically resemble the images on the left hand side, except for the last image that portrays an asian woman instead of a caucasian male.}
  \label{fig: UserStudySurgeon}
\end{figure*}
}
\end{document}